\documentclass[letterpaper, journal, final, twoside, dvipsnames, hidelinks]{IEEEtran}

\usepackage[T1]{fontenc}
\usepackage{amsmath,amsfonts}
\usepackage{algorithm}
\usepackage{algpseudocode}
\usepackage{array}
\usepackage[caption=false,font=normalsize,labelfont=sf,textfont=sf]{subfig}
\usepackage{textcomp}
\usepackage{stfloats}
\usepackage{url}
\usepackage{verbatim}
\usepackage{graphicx}
\usepackage{cite}
\usepackage{listofitems}
\usepackage{varwidth}
\usepackage{linegoal}

\usepackage{orcidlink}
\usetikzlibrary{plotmarks}
\usetikzlibrary{math}
\usetikzlibrary{calc}
\usetikzlibrary{decorations.pathreplacing}
\usepackage{pgfplots}
\pgfplotsset{compat=1.17}
\pgfplotsset{plot coordinates/math parser=false}
\usetikzlibrary{external}
\usetikzlibrary{shapes.geometric}
\usetikzlibrary{3d, perspective}
\usepgfplotslibrary{groupplots}
\usetikzlibrary{positioning}

\usepackage{calc}
\usepackage{lmodern}
\usepackage{anyfontsize}
\usepackage{mathrsfs}
\usepackage{cleveref}
\usepackage{url}
\usepackage{tabularray}
\usepackage{amssymb}

\usepackage{siunitx}

\usepackage[xindy]{glossaries}
\glsdisablehyper%

\usepackage{lscape}

\usepackage{placeins}

\usepackage{romannum}
\AtBeginDocument{\pagenumbering{arabic}}

\usepackage[htt]{hyphenat}
\renewcommand{\vec}[1]{\mathbf{\MakeLowercase{#1}}}
\newcommand{\Mat}[1]{\mathbf{\MakeUppercase{#1}}}
\newcommand{\set}[1]{\mathcal{#1}}
\newcommand{\intnum}{\mathbb{Z}}
\newcommand{\natnum}{\mathbb{N}}

\newcommand{\intmodring}[1]{\intnum_{#1}}
\newcommand{\entropy}{\mathscr{H}}
\newcommand{\nentropy}{\widetilde{\entropy}}
\newcommand{\Var}{\mathrm{Var}}
\newcommand{\nmse}{\ensuremath{\mathit{NMSE}}}
\newcommand{\mse}{\ensuremath{\mathit{MSE}}}
\newcommand{\primeset}{\mathscr{P}}
\newcommand{\subgroup}{\mathscr{S}}
\newcommand{\rulespace}{\mathcal{R}}

\usepackage{xargs}

\algrenewcommand\algorithmicrequire{\textbf{Input:}}
\algrenewcommand\algorithmicensure{\textbf{Output:}}
\algnewcommand\algorithmicforeach{\textbf{for each}}
\algdef{S}[FOR]{ForEach}[1]{\algorithmicforeach\ #1\ \algorithmicdo}
\newcommand{\Func}[1]{\textrm{#1}}
\newcommand{\Continue}{\textbf{continue}}
\algnewcommand{\LeftComment}[1]{\State{}\(\triangleright\) #1}

\crefformat{equation}{(#2#1#3)}
\crefrangeformat{equation}{(#3#1#4) to~(#5#2#6)}
\crefmultiformat{equation}{(#2#1#3)}%
{ and~(#2#1#3)}{, (#2#1#3)}{ and~(#2#1#3)}
\crefname{table}{Table}{Tables}
\Crefname{table}{Table}{Tables}
\crefname{figure}{Fig.}{Fig.}
\Crefname{figure}{Fig.}{Fig.}
\crefname{paragraph}{paragraph}{paragraphs}
\Crefname{paragraph}{Paragraph}{Paragraphs}
\Crefname{appsec}{appendix}{appendices}

\graphicspath{{figures/}}

\definecolor{tum-blue-brand}{RGB}{48, 112, 179}
\definecolor{tum-blue-dark}{RGB}{7, 33, 64}
\definecolor{tum-blue-light}{RGB}{94, 148, 212}
\definecolor{tum-blue-light-dark}{RGB}{154, 188, 228}
\definecolor{tum-blue-light-2}{RGB}{194, 215, 239}
\definecolor{tum-blue-light-3}{RGB}{215, 228, 244}
\definecolor{tum-blue-light-4}{RGB}{227, 238, 250}
\definecolor{tum-blue-light-5}{RGB}{240, 245, 250}
\definecolor{tum-yellow}{RGB}{254, 215, 2}
\definecolor{tum-yellow-2}{RGB}{254, 230, 103}
\definecolor{tum-yellow-dark}{RGB}{203, 171, 1}
\definecolor{tum-orange}{RGB}{247, 129, 30}
\definecolor{tum-orange-dark}{RGB}{217, 146, 8}
\definecolor{tum-pink}{RGB}{181, 92, 165}
\definecolor{tum-pink-dark}{RGB}{155, 70, 141}
\definecolor{tum-pink-1}{RGB}{198, 128, 187}
\definecolor{tum-pink-2}{RGB}{214, 164, 206}
\definecolor{tum-blue-bright}{RGB}{143, 129, 234}
\definecolor{tum-blue-bright-dark}{RGB}{105, 85, 226}
\definecolor{tum-red}{RGB}{234, 114, 55}
\definecolor{tum-red-dark}{RGB}{217, 81, 23}
\definecolor{tum-green}{RGB}{159, 186, 54}
\definecolor{tum-green-dark}{RGB}{125, 146, 42}
\definecolor{tum-grey-7}{RGB}{221, 226, 230}
\definecolor{tum-grey-8}{RGB}{235, 236, 239}

\definecolor{gradient-blue-1}{RGB}{94, 148, 212}
\definecolor{gradient-blue-2}{RGB}{114, 161, 184}
\definecolor{gradient-blue-3}{RGB}{133, 174, 155}
\definecolor{gradient-blue-4}{RGB}{153, 187, 127}

\newacronym{gcd}{GCD}{Greatest Common Divisor}
\newacronym{lcm}{LCM}{Least Common Multiple}
\newacronym[longplural={Frames per Second}]{fpsLabel}{FPS}{Frame per Second}
\newacronym{rc}{RC}{Reservoir Computing}
\newacronym[longplural={Cellular Automata}]{ca}{CA}{Cellular Automaton}
\newacronym{esn}{ESN}{Echo State Network}
\newacronym{dlr}{DLR}{Delay Line Reservoir}
\newacronym{scr}{SCR}{Simple Cycle Reservoir}
\newacronym{nn}{NN}{Neural Network}
\newacronym{lstm}{LSTM}{Long-Short Term Memory}
\newacronym{rnn}{RNN}{Recurrent Neural Network}
\newacronym{gru}{GRU}{Gated Recurrent Unit}
\newacronym{cnn}{CNN}{Convolutional Neural Network}
\newacronym{mg}{MG}{Mackey-Glass}
\newacronym{mg25}{MG\_25}{}
\newacronym{mso}{MSO}{Multiple Superimposed Oscillator}
\newacronym{mso3}{MSO\_3}{}
\newacronym{narma}{NARMA}{Nonlinear Autoregressive-Moving Average}
\newacronym{narma10}{NARMA\_10}{}
\newacronym{narma20}{NARMA\_20}{}
\newacronym{narma30}{NARMA\_30}{}
\newacronym{ncc}{NCC}{Nonlinear Communication Channel}
\newacronym{ppst}{PPST}{Pseudo Periodic Synthetic Time Series}
\newacronym{pmp}{PMP}{Predictive Modeling Problem}
\newacronym{relicada}{ReLiCADA}{Reservoir Computing using Linear Cellular Automata Design Algorithm}
\newacronym{reca}{ReCA}{Reservoir Computing using Cellular Automata}
\newacronym{relica}{ReLiCA}{Reservoir Computing using Linear Cellular Automata}
\newacronym{bpdc}{BPDC}{Backpropagation-Decorrelation}
\newacronym{elm}{ELM}{Extreme Learning Machine}
\newacronym{fpga}{FPGA}{Field-Programmable Gate Array}
\newacronym{ga}{GA}{Genetic Algorithm}
\newacronym{mse}{MSE}{Mean Squared Error}
\newacronym{nmse}{NMSE}{Normalized Mean Squared Error}
\newacronym{iid}{i.i.d.}{independent and identically distributed}
\newacronym{iot}{IoT}{Internet of Things}
\newacronym{msb}{MSB}{most significant bit}
\newacronym{eoc}{EoC}{Edge of Chaos}
\newacronym{eols}{EoLS}{Edge of Lyapunov Stability}
\newacronym{lsm}{LSM}{Liquid State Machine}

\newlength\figureheight%
\newlength\figurewidth%
\begin{document}

\bstctlcite{BSTcontrol}

\title{ReLiCADA - Reservoir Computing using Linear Cellular Automata Design Algorithm}

\author{Jonas~Kantic\,\orcidlink{0000-0001-5206-7790}, Fabian~C.~Legl\,\orcidlink{0000-0002-0214-7458}, Walter~Stechele\,\orcidlink{0000-0002-7455-8483}, Jakob~Hermann\,\orcidlink{0009-0001-3097-8697}
\thanks{This work was supported by the Bavarian Ministry of Economic Affairs,
Regional Development and Energy in the context of the Bavarian Collaborative Research Program (BayVFP), funding line Digitization, funding area Information, and Communication Technology.
\textit{(Jonas Kantic and Fabian Legl contributed equally to this work.) (Corresponding authors: Jonas Kantic and Fabian Legl.)}}%
\thanks{Jonas Kantic and Walter Stechele are with the Chair of Integrated Systems, Technical University of Munich (TUM), 80333 Munich, Germany (e-mail: jonas.kantic@tum.de, walter.stechele@tum.de).}%
\thanks{Fabian Legl and Jakob Hermann are with the Ingenieurb\"uro f\"ur Thermoakustik GmbH (IfTA GmbH), 82178 Puchheim, Germany (e-mail: fabian.legl@tum.de, jakob.hermann@ifta.com)}}

\markboth{}%
{}

\maketitle

\begin{abstract}
In this paper, we present a novel algorithm to optimize the design of \acrlong{reca} models for time series applications.
Besides selecting the models' hyperparameters, the proposed algorithm particularly solves the open problem of linear \acrlong{ca}
rule selection. The selection method pre-selects only a few promising candidate rules out of an exponentially growing rule space.
When applied to relevant benchmark datasets, the selected rules achieve low errors, with the best rules being among the top
\qty{5}{\percent} of the overall rule space.
The algorithm was developed based on mathematical analysis of linear \acrlong{ca} properties and is backed by almost one million experiments,
adding up to a computational runtime of nearly one year.
Comparisons to other state-of-the-art time series models show that the proposed \acrlong{reca} models have lower computational
complexity, at the same time, achieve lower errors.
Hence, our approach reduces the time needed for training and hyperparameter optimization by up to several orders of magnitude.
\end{abstract}


\glsresetall%

\section{Introduction}

\IEEEPARstart{R}{eal}-time sensor signal processing is a growing demand in our everyday life. High-frequent sensor data is available in a wide range of embedded applications, including, for example, speech recognition, battery protection in electric cars, and monitoring of production facilities.
Hence, many application domains could benefit from intelligent and real-time sensor signal analysis on low-cost and low-power devices.
To be able to adapt to changing environmental and operational conditions of the target system, machine learning-based approaches have to be employed since classical signal processing techniques often reach their limits under changing external influences. However, at the same time, efficient hardware implementation and acceleration of such intelligent methods are needed in order to fulfill real-time constraints for applications requiring inference times in the \unit{\micro{} \second} to \unit{\nano{} \second} range.

Currently, neural network-based models, including \glspl{rnn} and \glspl{lstm}, form the state-of-the-art for numerous time series processing tasks. However, these models typically consist of several different layers stacked in a deep architecture, have millions of trainable parameters, and include hard-to-implement nonlinear mappings. Such deep architectures are unfavorable for hardware implementations and induce long inference times, which is disadvantageous for real-time applications.

During the last two decades, \gls{rc} emerged as a promising alternative to deep and recurrent neural networks. In contrast to the latter, \gls{rc} models have a shallow architecture and trainable parameters in only a single layer.
This makes them generally much easier to design and train. Despite their comparatively simple architecture, \gls{rc} models have proven their capabilities in many application domains, like biomedical, audio, and social applications~\cite{Tanaka2019}.
While they only need relatively little computational resources compared to deep neural networks, they still have to be optimized for the unique requirements of, e.g., \gls{fpga}-based implementations.

Due to the discrete nature of their reservoirs, \gls{reca} models form a subset within the \gls{rc} framework that is suitable for the implementation on \glspl{fpga}. Like for other \gls{rc} models, the training of \gls{reca} models is easy and fast. Nevertheless, they require extensive hyperparameter tuning.

One major challenge that we address in this paper is that for most \gls{reca} models, the hyperparameter search space is too big for current heuristic search and optimization algorithms. This is especially true for the selection of suitable \gls{ca} rules in the reservoir.

Because of this, we conducted the first mathematical analysis of the influence of linear \gls{ca} rules on the model performance in the \gls{reca} framework and identified common analytical properties of suitable linear rules to be used in the reservoir.
We backed our mathematical analysis with the results of almost one million experiments with a sequential runtime of nearly one year using an NVIDIA RTX A4000 GPU (using this GPU, we were able to run three parallel runs of our experiment).
In the research community, the \gls{reca} framework has been tested almost solely on pathological datasets that do not allow conclusions about the generality of
the conducted studies and generalization capabilities of the \gls{reca} models themselves. In the context of this study, we performed an extensive analysis using several benchmark datasets.
The result of our research is the \gls{relicada}, which specifies \gls{relica} models with fixed hyperparameters, and thus immensely simplifies the overall design process.
The selected \gls{relica} models achieve lower errors than comparable state-of-the-art time series models while maintaining low computational complexity.

The rest of this paper is structured as follows.
We first start with an introduction to \gls{rc} and a review of related work in \cref{subsec:rc}, followed by \glspl{ca} in \cref{subsec:ca}, and finally,
an overview of the \gls{reca} framework in \cref{subsec:reca}. In \cref{subsec:math_params}, we define the mathematical parameters used in our analysis.
After that, we introduce our implementation and refinement of the \gls{reca} model architecture and describe all parts of it in \cref{sec:reca_proposed}.
This is followed by an explanation of our novel \acrlong{relicada} in \cref{sec:relicada}.
The datasets and models that we use to compare and validate our algorithm with are listed in \cref{sec:experiments}, before we analyze the experiments in \cref{sec:results}.
The paper is completed by a conclusion in \cref{sec:conclusion}.
\section{Background and Related Work}

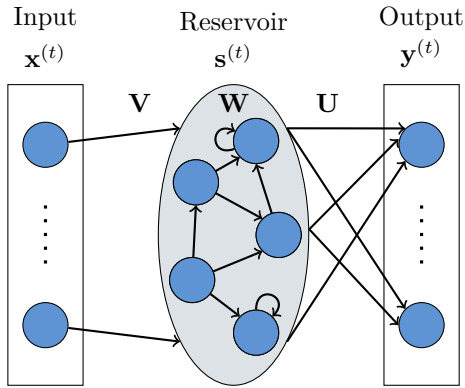
\begin{figure}[!t]
    \centering
    {
    \tikzstyle{esn-layer} = [rectangle, draw, fill=white, minimum width=1cm, minimum height=4cm]
    \tikzstyle{reservoir} = [ellipse, draw, fill=tum-grey-7, minimum width=2cm, minimum height=4cm]
    \tikzstyle{neuron} = [circle, draw, fill=tum-blue-light, minimum width=0.6cm, minimum height=0.6cm]

    \newcommand{\esnlayer}[3]{
        \setsepchar{,}
        \readlist\coordlist#2
        \begin{scope}[auto]
            \node[esn-layer] (#1) at (\coordlist[1]cm, \coordlist[2]cm) {};
            \node[neuron, below=0.5cm of #1.north] (#1n1) {};
            \node[neuron, above=0.5cm of #1.south] (#1n2) {};
            \draw[loosely dotted, very thick] ($(#1n1.south) - (0, 0.5cm)$) to ($(#1n2.north) + (0, 0.5cm)$);
            \node[align=center, above=0.1cm of #1] (#1cap) {#3};
        \end{scope}
    }

    \newcommand{\esnreservoir}[3]{
        \setsepchar{,}
        \readlist\coordlist#2
        \begin{scope}[auto]
            \coordinate (pos) at (\coordlist[1]cm, \coordlist[2]cm);
            \node[reservoir] (#1) at (pos) {};
            \node[neuron] (#1n1) at ($(pos) + (0.3cm, 1.25cm)$) {};
            \node[neuron] (#1n2) at ($(pos) + (-0.5cm, 0.7cm)$) {};
            \node[neuron] (#1n3) at ($(pos) + (0.6cm, 0.0cm)$) {};
            \node[neuron] (#1n4) at ($(pos) + (-0.55cm, -0.6cm)$) {};
            \node[neuron] (#1n5) at ($(pos) + (0.3cm, -1.3cm)$) {};

            \draw[->, thick] (#1n1) to [out=200, in=155, loop, looseness=4] (#1n1);
            \draw[->, thick] (#1n2.center) to (#1n1.south west);
            \draw[->, thick] (#1n2.center) to (#1n3.north west);
            \draw[->, thick] (#1n3.center) to (#1n1.south);
            \draw[->, thick] (#1n4.center) to (#1n2.south);
            \draw[->, thick] (#1n4.center) to (#1n5.north west);
            \draw[->, thick] (#1n4.center) to (#1n3.south west);
            \draw[->, thick] (#1n5) to [out=90, in=45, loop, looseness=4] (#1n5);

            \node[neuron] (#1n6) at ($(pos) + (0.3cm, 1.25cm)$) {};
            \node[neuron] (#1n7) at ($(pos) + (-0.5cm, 0.7cm)$) {};
            \node[neuron] (#1n8) at ($(pos) + (0.6cm, 0.0cm)$) {};
            \node[neuron] (#1n9) at ($(pos) + (-0.55cm, -0.6cm)$) {};
            \node[neuron] (#1n10) at ($(pos) + (0.3cm, -1.3cm)$) {};

            \node[align=center, above=0.1cm of #1] (#1cap) {#3};
        \end{scope}
    }

    \begin{tikzpicture}[node distance=0, auto]
        \esnlayer{input}{{0,0}}{\(\vec{x}^{(t)}\)}
        \esnreservoir{reservoir}{{2.5,0}}{\(\vec{s}^{(t)}\)}
        \esnlayer{output}{{5, 0}}{\(\vec{y}^{(t)}\)}
        \node[align=center, above=0.6cm of input] (input-text) {Input};
        \node[align=center, above=0.6cm of reservoir] (reservoir-text) {Reservoir};
        \node[align=center, above=0.6cm of output] (output-text) {Output};

        \draw[->, thick] (inputn1) to (reservoir.north west);
        \draw[->, thick] (inputn2) to (reservoir.south west);

        \draw[->, thick] (reservoir.north east) to (outputn1.north west);
        \draw[->, thick] (reservoir.mid east) to (outputn1.mid west);
        \draw[->, thick] (reservoir.south east) to (outputn1.south west);
        \draw[->, thick] (reservoir.mid east) to (outputn2.west);
        \draw[->, thick] (reservoir.north east) to (outputn2.north west);

        \node[align=center] (V) at ($(input.north) + (1.25, -0.25cm)$) {\(\Mat{V}\)};
        \node[align=center] (W) at ($(reservoir.north) - (0, 0.25cm)$) {\(\Mat{W}\)};
        \node[align=center] (U) at ($(output.north) - (1.25, 0.25cm)$) {\(\Mat{U}\)};

    \end{tikzpicture}
}
    \caption{Echo State Network as an example for Reservoir Computing.}%
    \label{fig:esn}
\end{figure}

The in-depth analysis of \gls{reca} models comprises concepts and methods from
different research fields, ranging from abstract algebra over automaton theory
and properties of dynamical systems to machine learning.
In the following sections, we summarize the required background knowledge and related work about \gls{rc}, \gls{ca}, and \gls{reca}.
Furthermore, we define the mathematical parameters that we use to characterize the \gls{reca} models.

\subsection{Reservoir Computing}%
\label{subsec:rc}

The main idea of \gls{rc} is to transform the input \({\vec{x}}\) into a higher dimensional space \({\vec{s}}\) in order to make the input
linearly separable.
This transformation is performed by a dynamic system which is called the reservoir (center part of \cref{fig:esn}).

The readout layer (right part of \cref{fig:esn}) is then used to linearly transform the reservoir state into the desired output
\({\vec{y}}\)~\cite{Nakajima2021}.
Generally, \gls{rc} models can be described using
\begin{equation}
    \begin{split}
        \vec{s}^{(t)} &= g(\Mat{V}\vec{x}^{(t)}, \Mat{W}\vec{s}^{(t-1)}) \\
        \vec{y}^{(t)} &= h(\Mat{U}\vec{s}^{(t)})
    \end{split}
\end{equation}
with the reservoir state \({\vec{s}}\), the input \({\vec{x}}\), and the output \({\vec{y}}\) at the discrete time \({t}\).
The function \({g}\) depends on the reservoir type, while the function \({h}\) describes the used readout layer and is typically a linear mapping.
During model training, only the output weights \({\Mat{U}}\) are trained, while the input weights \({\Mat{V}}\) and reservoir weights  \({\Mat{W}}\) are fixed and usually generated using some model specific constraints.
In \cref{fig:esn}, we depict an \gls{esn}~\cite{Jaeger2001} using a single layer \gls{rnn}~\cite{Rumelhart1986} as the reservoir.
Further simplifications to the reservoir were proposed by Rodan et al.~\cite{Rodan2011}, resulting in, e.g., the \gls{dlr} or the \gls{scr}. These types of reservoirs require less computations during the inference step compared to
general \glspl{esn}. Nevertheless, they are still not suited for implementation in, e.g., \glspl{fpga} due to the required floating point calculations.
To eliminate the floating point operations in the reservoir, stochastic bitstream neurons can be used~\cite{Verstraeten2005}.
However, stochastic bitstream neurons trade inference speed with the simplicity of implementation on \glspl{fpga} and are thus not suited for our usecase~\cite{Alomar2016}.
In this paper, we are focusing on a class of \gls{rc} models that use \glspl{ca} as the reservoir, which was first
proposed by Yilmaz~\cite{Yilmaz2014} and has been termed \gls{reca}~\cite{Margem2016}.
One of the main advantages of \gls{reca} models compared to other \gls{rc} models is that the reservoir only uses
integer operations on a finite set of possible values. Because of that, they are easy and fast to compute on digital systems like \glspl{fpga}.%
\footnote{There are a lot more models inside the \gls{rc} framework like \gls{lsm}~\cite{Natschlaeger2002}, \gls{elm}~\cite{Huang2006}, \gls{bpdc}~\cite{Steil2004}, and physical reservoirs~\cite{Tanaka2019}. Nonetheless, we will not go into details about them since they are not suitable for our target implementation.}
\subsection{Cellular Automata}%
\label{subsec:ca}

\begin{figure}[!t]
    \centering
    {
    \def\cxrad{3.0}
    \def\cyrad{3.0}
    \def\cheight{1.2}
    \definecolor{gradient-1}{RGB}{119, 165, 219}
    \definecolor{gradient-2}{RGB}{144, 182, 226}
    \definecolor{gradient-3}{RGB}{169, 199, 232}
    \def\cellgradone{tum-blue-light}
    \def\cellgradtwo{gradient-1}
    \def\cellgradthree{gradient-2}
    \def\cellgradfour{gradient-3}
    \def\cellgradfive{tum-blue-light-2}
    \def\cellbackcolor{tum-grey-7}
    \def\anglestep{20}
    \def\startangle{-40}

    \begin{tikzpicture}
        \begin{scope}[3d view={-40}{20}]
            \tikzmath{
                coordinate \c;
                \c = (0, 0);
                int \j;
                int \nextangle;
                \nextangle = \startangle + \anglestep;
                for \j in {\startangle, \nextangle, ..., 320}{
                    int \i;
                    \i = \j - \anglestep / 2;
                    int \k;
                    \k = \j - \anglestep;
                    {
                        \path[draw,fill=none, draw=none] (\c) arc[start angle=\k, end angle=\i, x radius=\cxrad, y radius=\cyrad] coordinate (c1\i) arc[start angle=\i, end angle=\j, x radius=\cxrad, y radius=\cyrad] coordinate (c1\j) -- ++(0,0,-\cheight) arc[start angle=\j, end angle=\i, x radius=\cxrad, y radius=\cyrad] coordinate (c2\i) arc[start angle=\i, end angle=\k, x radius=\cxrad, y radius=\cyrad] coordinate (c2\k) -- ++(0,0,\cheight);
                    };
                    \c = (c1\j);
                };
            }
            \tikzmath{
                real \mheight;
                \mheight = \cheight / 2;
                {
                    \path[draw, fill=tum-blue-light] ($(c2230) - (0,0,2.0)$)%
                    arc[start angle=230, end angle=240, x radius=\cxrad, y radius=\cyrad]%
                    coordinate (s0i1southeast) -- ++(0,0,0.67*\cheight)%
                    coordinate (s0i1east) -- ++(0,0,0.33*\cheight)%
                    coordinate (s0i1northeast) arc[start angle=240, end angle=230, x radius=\cxrad, y radius=\cyrad]%
                    coordinate (s0i1north) arc[start angle=230, end angle=220, x radius=\cxrad, y radius=\cyrad]%
                    coordinate (s0i1northwest) -- ++(0,0,-0.33*\cheight)%
                    coordinate (s0i1west) -- ++(0,0,-0.67*\cheight)%
                    arc[start angle=220, end angle=230, x radius=\cxrad, y radius=\cyrad];
                    \node[align=center, rotate=0] (s0i1) at ($(s0i1north) - (0,0,\mheight)$) {\(s_{0}^{(i)}\)};
                };
            }
            \draw[thick, ->, tum-red, text=black] (c2230) -- node[font=\small, fill=white, inner sep=1pt]{\(w_{0}\)} (s0i1north);
            \draw[thick, ->, tum-red, text=black] (c2250) -- node[font=\small, fill=white, inner sep=1pt]{\(w_{1}\)} (s0i1northeast);
            \draw[thick, ->, tum-red, text=black] (c2210) -- node[font=\small, fill=white, inner sep=1pt]{\(w_{-1}\)} (s0i1northwest);
            \draw[thick, ->, densely dotted, tum-green, text=black] (c2270) -- node[font=\small, right]{\(w_{2}\)} (s0i1east);
            \draw[thick, ->, densely dotted, tum-green, text=black] (c2190) -- node[font=\small, left]{\(w_{-2}\)} (s0i1west);
            \tikzmath{
                coordinate \c;
                \c = (0, 0);
                int \j;
                int \nextangle;
                \nextangle = \startangle + \anglestep;
                for \j in {\startangle, \nextangle, ..., 320}{
                    int \i;
                    \i = \j - \anglestep / 2;
                    int \k;
                    \k = \j - \anglestep;
                    \ccol = "\cellgradone";
                    if \j<160 then { \ccol = "\cellbackcolor"; };
                    if \j==220 || \j==260 then { \ccol = "\cellgradtwo"; };
                    if \j==200 || \j==280 then { \ccol = "\cellgradthree"; };
                    if \j==180 || \j==300 then { \ccol = "\cellgradfour"; };
                    if \j==160 || \j==320 then { \ccol = "\cellgradfive"; };
                    {
                        \path[draw,fill=\ccol] (\c) arc[start angle=\k, end angle=\i, x radius=\cxrad, y radius=\cyrad] arc[start angle=\i, end angle=\j, x radius=\cxrad, y radius=\cyrad] -- ++(0,0,-\cheight) arc[start angle=\j, end angle=\i, x radius=\cxrad, y radius=\cyrad] arc[start angle=\i, end angle=\k, x radius=\cxrad, y radius=\cyrad] -- ++(0,0,\cheight);
                    };
                    \c = (c1\j);
                };
            }
            \tikzmath{
                real \mheight;
                \mheight = \cheight / 2;
                {
                    \node[align=center, rotate=-30] (ld) at ($(c1170) - (0,0,\mheight)$) {\footnotesize \(\ldots\)};
                    \node[align=center, rotate=-4] (sn2) at ($(c1190) - (0,0,\mheight)$) {\scriptsize \(s_{N-2}^{(i-1)}\)};
                    \node[align=center, rotate=0] (sn1) at ($(c1210) - (0,0,\mheight)$) {\small \(s_{N-1}^{(i-1)}\)};
                    \node[align=center, rotate=0] (s0) at ($(c1230) - (0,0,\mheight)$) {\(s_{0}^{(i-1)}\)};
                    \node[align=center, rotate=0] (s1) at ($(c1250) - (0,0,\mheight)$) {\small \(s_{1}^{(i-1)}\)};
                    \node[align=center, rotate=4] (s2) at ($(c1270) - (0,0,\mheight)$) {\scriptsize \(s_{2}^{(i-1)}\)};
                    \node[align=center, rotate=30] (rd) at ($(c1290) - (0,0,\mheight)$) {\footnotesize \(\ldots\)};
                };
            }
        \end{scope}
    \end{tikzpicture}
}
    \caption{Lattice of a one-dimensional Cellular Automaton with periodic boundary conditions. Using only the orange weights results in \({n = 3}\); using the orange and green weights results in \({n = 5}\). The state of the cell \({s_0}\)
    in the \({(i)\textsuperscript{th}}\) iteration is the weighted sum of the cell states in its neighborhood in the \({(i-1)\textsuperscript{th}}\) iteration.}%
    \label{fig:ca-lattice}
\end{figure}
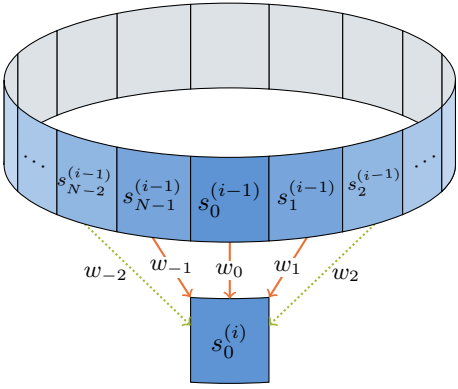

\Glspl{ca} represent one of the simplest types of time, space, and value discrete dynamical systems and have been
introduced initially by von Neumann~\cite{Neumann1963, Neumann1966}.
Following this idea, \glspl{ca} have been analyzed concerning several different properties,
including structural~\cite{Li1990, Wolfram2002}, algebraic~\cite{Ito1983, Wolfram1983, Martin1984, Das1992, Voorhees2012},
dynamical~\cite{Shereshevsky1992, Hurd1992, Kari1994, Damico2003, Burguet2021},
and behavioral~\cite{Codd1968, Packard1988, Langton1990, Mitchell1993, Teuscher2022} properties.

The \glspl{ca} considered in this paper consist of a finite, regular, and one-dimensional lattice of \({ N }\) cells (see \cref{fig:ca-lattice}), for reasons discussed later.
Each of the cells can be in one of \({ m }\) discrete states.
The lattice is assumed to be circularly closed, resulting in periodic boundary conditions.
In this sense, the right neighbor of the rightmost cell (\({s_{N-1}}\)) is the leftmost cell (\({s_0}\)), and vice versa.
A configuration of a \gls{ca} at a discrete
iteration\footnote{Since the time domain of the \gls{ca} is different from the time domain of the input data and \gls{reca} model, we will use iterations to describe the \gls{ca} evolution.}
\({ i }\) consists of
the states of all its cells at that iteration and can thus be written as a state vector \({ \vec{s}^{(i)} \in \intmodring{m}^{N} }\) according to
\begin{equation}
    \vec{s}^{(i)} = (s_0^{(i)}, \ldots, s_{N-1}^{(i)}) \text{, \qquad with } s_k \in \intmodring{m},
\end{equation}
where \({ \intmodring{m} = \intnum/m\intnum }\) denotes the ring of integers modulo \({ m }\), and \({(i)}\) denotes the iteration index.
The states of each cell change over the iterations according to a predefined rule. At iteration \({ (i) }\),
the cell state \({ s_k^{(i)} }\) is defined in dependency of the states of the cells
in its neighborhood of fixed size \({ n }\) at iteration \({ (i-1) }\) (see \cref{fig:ca-lattice}). The neighborhood of a cell contains the cell itself,
as well as a range of \({ r }\) neighboring cells to the left and right, respectively, leading to
\begin{equation}
    n = 2r+1 \text{,\qquad with } r \in \natnum^+.
\end{equation}

We introduce a restriction to the neighborhood \({n}\) to define the true neighborhood \({\hat{n}}\). For \({\hat{n}}\) we require that \({w_{-r} \neq 0}\) or \({w_r \neq 0}\).

The iterative update of the cell states can be described in terms of a local rule
\({ f\colon \intmodring{m}^n \rightarrow \intmodring{m} }\), which defines the dynamic behavior of the \gls{ca} according to
\begin{equation}
    \label{eq:ca_local_rule}
    s_k^{(i)} = f(s_{k-r}^{(i-1)}, \ldots, s_{k+r}^{(i-1)}).
\end{equation}
Since we use periodic boundary conditions, the indices \({k-r, \ldots, k+r}\) of the states in \cref{eq:ca_local_rule} have to be taken \({\text{mod } N}\).

Linear \glspl{ca} form a subset of general \glspl{ca}~\cite{Voorhees2012}.
The local rule of a linear \gls{ca} is a linear combination of the cell states in the considered neighborhood. Hence,
for linear \glspl{ca}, \({ f }\) (see \cref{eq:ca_local_rule}) can be defined as
\begin{equation}
    \label{eq:ca_linear_rule}
    f(s_{k-r}^{(i)},\ldots, s_{k+r}^{(i)}) = \sum\limits_{j=-r}^{r}{w_j}s_{k+j}^{(i)}
\end{equation}
with rule coefficients \({ w_j \in \intnum_m}\). A linear rule can thus be identified by its rule coefficients \({\vec{w} = (w_{-r},\ldots,w_r)}\).
Unless otherwise noted, we will restrict the \gls{ca} rule to linear rules in this paper.
A prominent example is the elementary rule 90 \gls{ca}, which is defined by \({m=2}\), \({n = 3}\) and \({(w_{-1}, w_0, w_1) = (1,0,1)}\)~\cite{Wolfram1983}.

For each linear rule \({f}\), there exists a mirrored rule \({\hat{f}}\) with \({\hat{\vec{w}} = (w_r,\ldots,w_{-r})}\).
If the rule coefficients are symmetric with respect to the
central coefficient \({w_0}\), it holds that \({\hat{f} = f}\).
In total, there exist \({ m^n }\) different linear \gls{ca} rules, which directly follows from \cref{eq:ca_linear_rule}.
We denote the set of all linear rules for given \({m}\) and \({n}\) by
\begin{equation}
    \rulespace(m,n) = \{\left(w_{-r},\ldots,w_{r}\right) : w_i \in \intnum_m, n=2r+1\}.
    \label{eq:rule_space}
\end{equation}

The local rule \({ f }\) is applied simultaneously to every cell of the lattice, such that the configuration \({ \vec{s}^{(i-1)} }\) updates to the next iteration
\({ \vec{s}^{(i)} }\), and therefore it induces a global rule \({ F\colon \intmodring{m}^N \rightarrow \intmodring{m}^N }\).
For linear \glspl{ca}, this mapping of configurations can be described by multiplication with a circulant matrix \({ \Mat{W} \in \intmodring{m}^{N \times N} }\), which is given by
\begin{equation}
    \Mat{W} = \text{circ}(w_0, \ldots, w_r, 0, \ldots, 0, w_{-r}, \ldots, w_{-1}),
    \label{eq:circulant_mat}
\end{equation}
with \({ \text{circ} }\) as defined in~\cite{Voorhees2012} (\emph{note}: if \({ N = n }\), the circulant matrix has no additional zero entries that are not rule coefficients).
Thus, the global rule for a linear \gls{ca} can be written as
\begin{equation}
    \vec{s}^{(i)} = F(\vec{s}^{(i-1)}) = \Mat{W}\vec{s}^{(i-1)}.
\end{equation}

It has been shown that several key properties typically used to characterize dynamical systems
are not computable for general \glspl{ca}. Even for general one-dimensional \glspl{ca}
nilpotency is undecidable, and the topological entropy cannot even be approximated~\cite{Kari1992, Damico2003, Hurd1992}. Furthermore,
injectivity and surjectivity can be computed only for one- and two-dimensional \glspl{ca}~\cite{Kari1994, Damico2003}.
However, when restricting the analysis to one-dimensional linear \glspl{ca}, all the mentioned properties are computable.
This is the reason why we focus on one-dimensional linear \glspl{ca} in this paper.
\subsection{ReCA Framework}%
\label{subsec:reca}

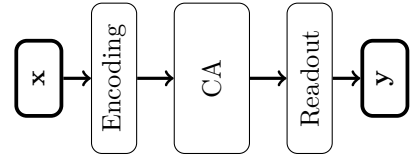
\begin{figure}[!t]
    \centering
    {
    \tikzstyle{inout} = [rectangle, draw, fill=white, rounded corners, minimum height=1cm, minimum width=0.6cm, line width=0.5mm]
    \tikzstyle{layer} = [rectangle, draw, fill=white, rounded corners, minimum height=2cm, minimum width=0.6cm]
    \tikzstyle{ca} = [rectangle, draw, fill=white, rounded corners, minimum height=2cm, minimum width=1cm]
    
    \begin{tikzpicture} [node distance = 1cm, auto]
        \node [inout] (in) {};
        \node[align=center,rotate=90] at (in.center) {\(\vec{x}\)};
        \node [layer, right of=in] (enc) {};
        \node[align=center,rotate=90] at (enc.center) {Encoding};
        \node [ca, right of=enc, xshift=3mm] (ca) {};
        \node[align=center,rotate=90] at (ca.center) {CA};
        \node [layer, right of=ca, xshift=3mm] (read) {};
        \node[align=center,rotate=90] at (read.center) {Readout};
        \node [inout, right of=read] (out) {};
        \node[align=center,rotate=90] at (out.center) {\(\vec{y}\)};
        \draw [->, very thick] (in) -- (enc);
        \draw [->, very thick] (enc) -- (ca);
        \draw [->, very thick] (ca) -- (read);
        \draw [->, very thick] (read) -- (out);
    \end{tikzpicture}
}
    \caption{\Gls{reca} architecture as initially proposed by Yilmaz~\cite{Yilmaz2014}}%
    \label{fig:reca_original}
\end{figure}

\glspl{ca} have been employed as the reservoir in the \gls{rc} framework first by Yilmaz~\cite{Yilmaz2014},
replacing the recurrently connected neurons typically used in \glspl{esn}.
The original architecture of a \gls{reca} model is depicted in \cref{fig:reca_original}.
Input to the model is a time series \({ \vec{x} }\), which is fed sample by sample into an encoding stage.
The encoding stage, as proposed in~\cite{Yilmaz2014}, serves several purposes. First, the input is preprocessed depending on
the type of data, which may include feature expansion, weighted summation, scaling, and binarization.
Second, the processed data is mapped to the cells of the CA in the reservoir. Third, the processed data is
encoded into the mapped cell states~\cite{Yilmaz2014}.
With the input encoded into its cell states, the global rule of the \gls{ca} is executed iteratively for a fixed number of iterations.
The output of the \gls{ca} is then passed to the readout layer, which produces the final model output \({\vec{y}}\).

The \gls{reca} framework has been analyzed and developed further based on the initially proposed architecture.
In Nichele et al.~\cite{Nichele2017a}, the authors use hybrid \gls{ca}-based reservoirs split into two halves,
each half running with a different rule to enrich the dynamics within the reservoir.
However, this increases the search space for suitable rule combinations in the reservoir, and it remains unclear how to design the reservoir effectively.
Deep reservoir computing using the \gls{reca} approach is investigated by Nichele et al.~\cite{Nichele2017} by stacking two \gls{reca} models one after the other,
resulting in decreased error rates in most of the analyzed cases.
This design principle, however, is to some point contradictory to the original intention of \gls{rc},
which is to reduce the complexity of supervised training of \glspl{nn}~\cite{Jaeger2002a}.
The analysis of suitable \gls{ca} rules has been extended from elementary \glspl{ca} (\({ m=2, \hat{n} = 3 }\)) to complex \glspl{ca}
(\({ m \ge 3}\) and/or \({\hat{n} \ge 5 }\)) in~\cite{Babson2019}. In their work, the authors use a \gls{ga} to perform a heuristic optimization within the super-exponentially growing rule space
(\({ m^{m^{\hat{n}}} }\), since they do not restrict on linear \glspl{ca}) to find suitable rules for use in the reservoir. One of the biggest challenges with this approach is that the rule space quickly becomes
unmanageable for heuristic optimization methods, including genetic algorithms. Even when the number of possible states is only doubled from, e.g., \({ m=2 }\) to only \({ m=4 }\), the
number of possible rules with a three-neighborhood grows from \({2^{2^{3}} = 256}\) to \({ 4^{4^{3}} \approx 3.4 \times 10^{38} }\).
This example impressively shows that even small increases in complexity of the
\gls{ca} reservoirs make applying heuristic search and optimization methods practically impossible.

Most of the research mentioned above has been mainly based on the synthetic 5-bit and 20-bit memory tasks~\cite{Yilmaz2014}.
However, as the authors in~\cite{Margem2020} point out, especially the 5-bit memory task is not sufficient to make conclusions
about the generalization capability of a model since this task consists of only 32 examples. Furthermore, the
model is trained and tested on the whole dataset which contradicts common practice of separating training and test sets. Therefore, they adapt the 5-bit memory task by splitting the 32 examples into a training and test set.
This, however, shrinks the number of available train and test examples further.
The authors also investigate the effect of different feature extraction techniques on the reservoir output with the result that simply overwriting \gls{ca} cells in the reservoir works well in less complex \glspl{ca}.

A rule 90-based \gls{fpga} implementation of a \gls{reca} model for the application of handwritten digit recognition based on the MNIST dataset is presented in~\cite{Moran2020}.
Even though their implementation does not reach the classification accuracy of current
state-of-the-art \gls{cnn}-based implementations, the authors show that \gls{reca} is a promising alternative to traditional neural network-based machine learning approaches.
This is especially underlined by the fact that the energy efficiency of their implementation is improved
by a factor of 15 compared to \gls{cnn} implementations~\cite{Moran2020}.

An analysis of the influence of several hyperparameters in the \gls{reca} framework has been conducted in~\cite{Glover2021}, with the result
that for general \glspl{ca}, the overall performance of the model is dependent on and sensitive to the concrete choice of hyperparameters.
\subsection{Mathematical Parameters}%
\label{subsec:math_params}

This section introduces the mathematical parameters we use to analyze linear \gls{ca} rules.
Depending on \({m}\), \({\intnum_m}\) is a finite field if \({m}\) is prime or otherwise a finite ring.
This has several mathematical effects, e.g., the existence of unique multiplicative inverses.
Unless otherwise noted, we assume the more general case where \({m}\) is not prime (\({\intnum_{m}}\) is a ring).

We define the prime factor decomposition of \({m}\) as
\begin{equation}
    m=p_1^{k_1} \cdots p_h^{k_h}
    \label{eq:prime_decomposition}
\end{equation}
with the set of prime factors as
\begin{equation}
     \primeset = \{p_1, \ldots, p_h\}
     \label{eq:prime_set}
\end{equation}
and their multiplicities
\begin{equation}
    \set{K} = \{k_1,\ldots,k_h\}.
    \label{eq:prime_multiplicities}
\end{equation}
The set of prime weights can be generated using
\begin{equation}
    \primeset_w = \{s:\gcd(s,m)=1\} \qquad \forall s \in \intnum_m {\setminus} 0
    \label{eq:prime_weights}
\end{equation}
and the set of non-prime weights by using
\begin{equation}
    \bar{\primeset}_w = \{s:\gcd(s,m) \neq 1\} \qquad \forall s \in \intnum_m {\setminus} 0
    \label{eq:non_prime_weights}
\end{equation}
where \textrm{gcd} denotes the greatest common divisor.

\subsubsection{Transient and Cycle Lengths}
The behavior of a \gls{ca} over time can be separated into a transient phase of length \({k}\) and a cyclic phase of length \({c}\).
For linear \glspl{ca}, this can be expressed as
\begin{equation}
    \Mat{W}^{k}\vec{s}^{(0)} = \Mat{W}^{k+c}\vec{s}^{(0)}
\end{equation}
with the circulant rule matrix \({\Mat{W}}\) and the initial configuration \({\vec{s}^{(0)}}\)~\cite{Martin1984,Mendivil2012}.
The decomposition of the state space of a \gls{ca} into transients and cycles gives further information about its dynamic behavior.
A linear \gls{ca} with no transient phase has no Garden-of-Eden states.
Garden-of-Eden states have no predecessors and can thus only appear as initial states, if the \gls{ca} has a transient phase. On the computation of transient and cycle lengths, we refer the interested reader to~\cite{Jen1988, Stevens1999, Sutner2002, Mendivil2012, Zhang2013, Qureshi2015, Qureshi2019}%
\footnote{We would like to thank C. Qureshi for his valuable input on the computation of cycle lengths of linear mappings over finite fields.}.

\subsubsection{Cyclic Subgroup Generation}
A cyclic subgroup is generated by a generator element \({g}\). This generator can be used to generate the multiplicative
\begin{equation}
    \subgroup^{\times}(g) = \{g^0, g^1, g^2, \ldots, g^{(m-1)}\}
\end{equation}
and additive
\begin{equation}
    \subgroup^{+}(g) = \{0, g, 2g, \ldots, (m-1)g\}
\end{equation}
cyclic subgroups~\cite{Lidl1996,Dummit1991,Forney2005}.

The order of the cyclic additive subgroup \({|\subgroup^+(g)|}\) can be calculated by
\begin{equation}
    |\subgroup^+(g)| = \frac{m}{\gcd(m,g)}
\end{equation}
\cite{Forney2005,Dummit1991}. We use the order of cyclic subgroups to analyze whether the set of possible states during an iteration of a linear \gls{ca} shrinks or not.

\subsubsection{Topological Properties}%
\label{subsec:topological_properties}
For a mathematical analysis, it is often convenient to consider infinite linear \glspl{ca} whose lattice consist of infinitely many cells~\cite{Wolfram1986}.
Hence, further properties of infinite one-dimensional \glspl{ca} can be defined that characterize the behavior of the \gls{ca} as a dynamic system.
For some properties, we only give informal and intuitive descriptions. Formal definitions can be found in~\cite{Manzini1999a, Cattaneo2000a, Damico2003, Butt2014}.
In the following, the symbol \(\exists _n\) denotes ``there exist exactly \(n\)-times''.

\begin{figure*}[!t]
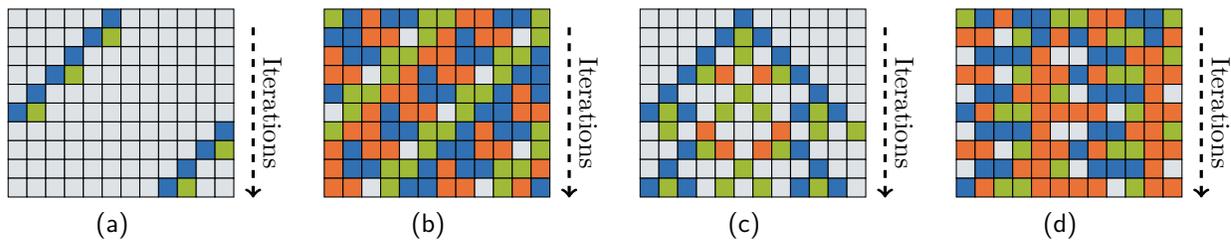

    \centering
    \hspace*{-2em}
    \subfloat[]{%
        {
    \input{figures/ca_rule_definitions}

    \begin{tikzpicture}
        \caconfig{r1}{{0,-0.0}}{{0,0,0,0,0,1,0,0,0,0,0,0}}
        \caconfig{r2}{{0,-0.25}}{{0,0,0,0,1,2,0,0,0,0,0,0}}
        \caconfig{r3}{{0,-0.5}}{{0,0,0,1,0,0,0,0,0,0,0,0}}
        \caconfig{r4}{{0,-0.75}}{{0,0,1,2,0,0,0,0,0,0,0,0}}
        \caconfig{r5}{{0,-1.0}}{{0,1,0,0,0,0,0,0,0,0,0,0}}
        \caconfig{r6}{{0,-1.25}}{{1,2,0,0,0,0,0,0,0,0,0,0}}
        \caconfig{r7}{{0,-1.5}}{{0,0,0,0,0,0,0,0,0,0,0,1}}
        \caconfig{r8}{{0,-1.75}}{{0,0,0,0,0,0,0,0,0,0,1,2}}
        \caconfig{r9}{{0,-2.0}}{{0,0,0,0,0,0,0,0,0,1,0,0}}
        \caconfig{r10}{{0,-2.25}}{{0,0,0,0,0,0,0,0,1,2,0,0}}
        \draw[->, very thick, dashed] ($(r2c12.north east) + (0.25,0)$) -- node[right]{\rotatebox{-90}{Iterations}} ($(r10c12.south east) + (0.25,0)$);
    \end{tikzpicture}
}\label{fig:rule_impulse}
    }%
    \hspace*{-4em}
    \subfloat[]{%
        {
    \input{figures/ca_rule_definitions}

    \begin{tikzpicture}
        \caconfig{r1}{{7,-0.0}}{{2,1,3,1,1,2,2,3,3,1,1,2}}
        \caconfig{r2}{{7,-0.25}}{{1,1,3,3,0,2,3,1,3,3,0,2}}
        \caconfig{r3}{{7,-0.5}}{{3,1,1,2,2,3,3,1,1,2,2,1}}
        \caconfig{r4}{{7,-0.75}}{{3,3,0,2,3,1,3,3,0,2,1,1}}
        \caconfig{r5}{{7,-1.0}}{{1,2,2,3,3,1,1,2,2,1,3,1}}
        \caconfig{r6}{{7,-1.25}}{{0,2,3,1,3,3,0,2,1,1,3,3}}
        \caconfig{r7}{{7,-1.5}}{{2,3,3,1,1,2,2,1,3,1,1,2}}
        \caconfig{r8}{{7,-1.75}}{{3,1,3,3,0,2,1,1,3,3,0,2}}
        \caconfig{r9}{{7,-2.0}}{{3,1,1,2,2,1,3,1,1,2,2,3}}
        \caconfig{r10}{{7,-2.25}}{{3,3,0,2,1,1,3,3,0,2,3,1}}
        \draw[->, very thick, dashed] ($(r2c12.north east) + (0.25,0)$) -- node[right]{\rotatebox{-90}{Iterations}} ($(r10c12.south east) + (0.25,0)$);
    \end{tikzpicture}
}\label{fig:rule_random}
    }%
    \hspace*{-4em}
    \subfloat[]{%
        {
    \input{figures/ca_rule_definitions}

    \begin{tikzpicture}
        \caconfig{r1}{{7,-0.0}}{{0,0,0,0,0,1,0,0,0,0,0,0}}
        \caconfig{r2}{{7,-0.25}}{{0,0,0,0,1,2,1,0,0,0,0,0}}
        \caconfig{r3}{{7,-0.5}}{{0,0,0,1,0,2,0,1,0,0,0,0}}
        \caconfig{r4}{{7,-0.75}}{{0,0,1,2,3,0,3,2,1,0,0,0}}
        \caconfig{r5}{{7,-1.0}}{{0,1,0,0,0,2,0,0,0,1,0,0}}
        \caconfig{r6}{{7,-1.25}}{{1,2,1,0,2,0,2,0,1,2,1,0}}
        \caconfig{r7}{{7,-1.5}}{{0,2,0,3,0,0,0,3,0,2,0,2}}
        \caconfig{r8}{{7,-1.75}}{{0,0,1,2,3,0,3,2,1,0,0,0}}
        \caconfig{r9}{{7,-2.0}}{{0,1,0,0,0,2,0,0,0,1,0,0}}
        \caconfig{r10}{{7,-2.25}}{{1,2,1,0,2,0,2,0,1,2,1,0}}
        \draw[->, very thick, dashed] ($(r2c12.north east) + (0.25,0)$) -- node[right]{\rotatebox{-90}{Iterations}} ($(r10c12.south east) + (0.25,0)$);
    \end{tikzpicture}
}\label{fig:rule_impulse_hign_ent}
    }%
    \hspace*{-4em}
    \subfloat[]{%
        {
    \input{figures/ca_rule_definitions}

    \begin{tikzpicture}
        \caconfig{r1}{{7,-0.0}}{{2,1,3,1,1,2,2,3,3,1,1,2}}
        \caconfig{r2}{{7,-0.25}}{{3,3,0,2,1,3,1,3,2,2,1,3}}
        \caconfig{r3}{{7,-0.5}}{{0,1,1,1,3,0,0,1,1,3,3,2}}
        \caconfig{r4}{{7,-0.75}}{{3,3,0,2,3,3,1,3,2,2,3,3}}
        \caconfig{r5}{{7,-1.0}}{{0,1,1,3,3,2,0,1,1,1,3,0}}
        \caconfig{r6}{{7,-1.25}}{{1,3,2,2,3,3,3,3,0,2,3,3}}
        \caconfig{r7}{{7,-1.5}}{{0,1,1,1,3,0,0,1,1,3,3,2}}
        \caconfig{r8}{{7,-1.75}}{{3,3,0,2,3,3,1,3,2,2,3,3}}
        \caconfig{r9}{{7,-2.0}}{{0,1,1,3,3,2,0,1,1,1,3,0}}
        \caconfig{r10}{{7,-2.25}}{{1,3,2,2,3,3,3,3,0,2,3,3}}
        \draw[->, very thick, dashed] ($(r2c12.north east) + (0.25,0)$) -- node[right]{\rotatebox{-90}{Iterations}} ($(r10c12.south east) + (0.25,0)$);
    \end{tikzpicture}
}\label{fig:rule_random_high_ent}
    }%
    \caption{Iteration diagram of linear \gls{ca} with \({m=4}\), \({\hat{n}=3}\), \({N=12}\), \({\vec{w} = (0,2,1)}\) (resulting in \({H=2}\)) and (a) a single cell initialized with state 1 (impulse) or (b) random initial configuration for \({I=9}\) iterations. Figures (c) and (d) have the same setup, but with \({\vec{w} = (1,2,1)}\) (resulting in \({H=4}\)). The colors indicate different cell states in \({\intnum_m}\).}%
    \label{fig:rule_iteration}
\end{figure*}

\paragraph{State Space and Orbit}
Intuitively, the set of all possible lattice configurations for infinite \glspl{ca} can be thought of as forming a state space.
Furthermore, the notion of distance that induces a metric topology on the state space can be integrated. For a detailed definition, we refer the interested reader to~\cite{Manzini1999a}.
An individual element in this set is a specific state configuration of the lattice.
The series of points in the state space during operation of an infinite linear \gls{ca}, i.e., the path \({(\vec{s}^{(0)}, \dots, \vec{s}^{(I)})}\) along the visited lattice configurations under iteration of \({F}\) for \({I}\) iterations with initial configuration \({\vec{s}^{(0)}}\), is called orbit.
Based on this topological framework, further properties of the dynamic behavior of linear one-dimensional \glspl{ca} can be computed that characterize
it for the asymptotic case of \({N \rightarrow \infty}\). However, only finite lattices can be realized in practical implementations and simulations of \glspl{ca},
whereby periodic boundary conditions have only limited influence on the behavior of the \gls{ca} compared to static
boundary conditions~\cite{LuValle2019}.

\paragraph{Topological Entropy}
The topological entropy is a measure of uncertainty of a dynamical system under repeated application of its mapping function (global
rule \({F}\) for infinite linear \glspl{ca}) starting with a partially defined initial state~\cite{Damico2003}. It can be used to characterize the asymptotic behavior of the system with respect to its operation.
Since discrete and finite dynamical systems fall into periodic state patterns, the topological entropy gives an idea of the complexity of the orbit structure and can be used to distinguish ordered and chaotic dynamical systems. For example, two runs of the same (infinite) linear \gls{ca} with different initial configurations that are close in the state space can be considered.
If the linear \gls{ca} has a low entropy, the final states of the two runs are also likely to be close in the state space~\cite{Butt2014}. However, suppose the \gls{ca} has a high topological entropy. In that case,
it shows chaotic behavior and the \gls{ca} is likely to produce diverging orbits during the two runs even though the initial states were close. Hence, a high entropy leads to increased uncertainty in the dynamical system's behavior.
This behavior can also be seen in \cref{fig:rule_iteration}, where the orbits of the rule with smaller entropy (\cref{fig:rule_impulse,fig:rule_random}) show a less chaotic behavior compared to the orbits of the rule with higher entropy (\cref{fig:rule_impulse_hign_ent,fig:rule_random_high_ent}).

The topological entropy (probabilistic approach) is closely related to the Lyapunov exponents (geometric approach) and can be computed based thereon. Assuming a CA over \({\intnum_m}\),
with the prime factor decomposition \cref{eq:prime_decomposition}, we define for \({i=1, \ldots, h}\)
\begin{equation}
    \begin{split}
        \set{P}_i &= \left\{ 0 \right\} \cup \left\{ j: \gcd\left( w_j, p_i \right) = 1 \right\}\\
        L_i &= \min \set{P}_i \\
        R_i &= \max \set{P}_i
    \end{split}
\end{equation}
with \({w_j}\) as defined in \cref{eq:ca_linear_rule}. Then the left \({\lambda^{-}}\) and right \({\lambda^{+}}\) Lyapunov exponents are~\cite{Damico2003}
\begin{equation}
    \begin{split}
        \lambda^{-} &= \max_{1 \leq i \leq h}\left\{ R_i \right\}\\
        \lambda^{+} &= -\min_{1 \leq i \leq h}\left\{ L_i \right\}.
    \end{split}
\end{equation}
The topological entropy can be calculated using~\cite{Damico2003}
\begin{equation}
    \entropy = \sum_{i=1}^{h} k_i \left( R_i - L_i \right)\log_2\left( p_i \right).
\end{equation}
To be able to compare the topological entropy of a \gls{ca} acting on different-sized finite rings, we introduce the normalized topological entropy
\begin{equation}
    \nentropy = \frac{\entropy}{\sum_{i=1}^{h}k_i\log_2\left( p_i \right)} = \frac{\entropy}{\log_2\left( m \right)}
    \label{eq:normalized_entropy}
\end{equation}
with \({m}\) as defined in \cref{eq:prime_decomposition}. For prime power rings, \({\nentropy}\) will only be integer values, where \({\nentropy = 1}\) will be the
smallest nonzero entropy, \({\nentropy = 2}\) the second smallest etc.

\paragraph{Equicontinuity}
A linear \gls{ca} is said to be \textit{equicontinuous} (or \textit{stable}) if any two states within a fixed size neighborhood in the state space diverge by at most some upper bound distance under iteration of \({F}\)~\cite{Manzini1999a}.
\textit{Equicontinuity} is given if the linear \gls{ca} fulfills the condition~\cite{Manzini1999a}
\begin{equation}
    \label{eq:equicontinuity}
    (\forall p \in \primeset) : p \mid \gcd(m, w_{-r}, \ldots, w_{-1}, w_1, \ldots, w_r).
\end{equation}

\paragraph{Sensitivity}
On the other hand, a linear \gls{ca} is \textit{sensitive} to initial conditions if, for any initial state \({\vec{s}^{(0)}}\), there exists another distinct initial state in any arbitrarily small neighborhood of \({\vec{s}^{(0)}}\), such that both orbits diverge by at least some lower bound distance~\cite{Manzini1999a}.
If the condition
\begin{equation}
    \label{eq:sensitivity}
    (\exists p \in \primeset) : p \nmid \gcd(m, w_{-r}, \ldots, w_{-1}, w_1, \ldots, w_r)
\end{equation}
is fulfilled, the corresponding \gls{ca} is \textit{sensitive}~\cite{Manzini1999a}.

\paragraph{Expansivity}
Suppose the orbits of any two different states in the state space diverge by at least some lower bound distance under forward iteration of \({F}\). In that case, the corresponding \gls{ca} is called \textit{positively expansive}~\cite{Manzini1999a}. Compared to \textit{sensitivity}, \textit{positive expansivity} is a stronger property.
\textit{Positive expansivity} is given for a linear \gls{ca} if~\cite{Manzini1999a}
\begin{equation}
    \label{eq:positive-expansive}
    \gcd(m, w_{-r}, \ldots, w_{-1})=\gcd(m, w_{1}, \ldots, w_{r})=1.
\end{equation}

For invertible infinite linear \glspl{ca}, this concept can be generalized by additionally considering backward iteration of \({F}\) and calling such \glspl{ca} \textit{expansive}~\cite{Manzini1999a}.
The condition for \textit{expansivity} is the same as \cref{eq:transitivity} for linear \glspl{ca}.

\paragraph{Transitivity}
\textit{Transitivity} is given for a linear \gls{ca}, if it has states that eventually move under iteration of \({F}\) from one arbitrarily small neighborhood to any other~\cite{Cattaneo2000a}. In other words, the linear \gls{ca} cannot be divided into independent subsystems. Codenotti and Margara~\cite{Codenotti1996} showed that, for \glspl{ca}, \textit{transitivity} implies \textit{sensitivity}.
The condition for \textit{transitivity} of a linear \gls{ca} is~\cite{Cattaneo2000a}
\begin{equation}
    \label{eq:transitivity}
    \gcd(m, w_{-r}, \ldots, w_{-1}, w_1, \ldots, w_r)=1.
\end{equation}

In addition, \textit{strong transitivity} is given if a \gls{ca} has orbits that include every state of its state space.
For \textit{strong transitivity}, a linear \gls{ca} must fulfill the condition~\cite{Manzini1999a}
\begin{equation}
    \label{eq:strong-transitivity}
    (\forall p \in \primeset)(\exists w_i,w_j) : p \nmid w_i \land p \nmid w_j.
\end{equation}

\paragraph{Ergodicity}
In contrast to \textit{transitivity}, \textit{ergodicity} concerns statistical properties of the orbits of a dynamical system.
While \textit{transitivity} indicates the state space of infinite linear \glspl{ca} cannot be separated, \textit{ergodicity}, intuitively, denotes the fact that typical orbits of almost all initial states (except for a set of points with measure zero) in any subspace under iteration of \({F}\) eventually revisit the entire set with respect to the normalized Haar measure~\cite{Cattaneo1997a, Sato1997}.
Cattaneo et al.~\cite{Cattaneo1997a} show that, for infinite linear \glspl{ca}, \textit{ergodicity} and \textit{transitivity} are equivalent.
The condition for a linear \gls{ca} to be \textit{ergodic} is the same as \cref{eq:transitivity}.

\paragraph{Regularity}
If cyclic orbits are dense in the state space for an infinite linear \gls{ca}, then it is denoted as \textit{regular}~\cite{Cattaneo2000a}.
\textit{Regularity} is defined for linear \gls{ca} by condition~\cite{Cattaneo2000a}
\begin{equation}
    \label{eq:regular}
    \gcd(m, w_{-r}, \ldots, w_r)=1.
\end{equation}

\paragraph{Surjectivity and Injectivity}
The global rule \(F\) of a linear \gls{ca} is \textit{surjective} if every state configuration has a predecessor. Thus, \textit{surjective} \glspl{ca} have no Garden-of-Eden
states and no transient phase~\cite{Voorhees2012}. Cattaneo et al.~\cite{Cattaneo1997a} showed that transitive \glspl{ca} are surjective. For one-dimensional \glspl{ca}, \textit{surjectivity} is
equivalent to \textit{regularity} of the global rule \({F}\)~\cite{Cattaneo2000a}.\ \textit{Surjectivity} for \(F\) is
given if condition \cref{eq:regular} is fulfilled~\cite{Ito1983}.

\textit{Injectivity} of \(F\) denotes the fact that every state has at most one predecessor. Every \textit{injective} \gls{ca} is also \textit{surjective}~\cite{Manzini1999a}.
If \(F\) is \textit{surjective} and \textit{injective}, the \gls{ca} is called \textit{bijective}, which is equivalent to reversibility~\cite{Voorhees2012}.
The condition for \textit{injectivity} of a linear \gls{ca} is given by~\cite{Ito1983}
\begin{equation}
    \label{eq:injective}
    (\forall p \in \mathscr{P})(\exists_1 w_i) : p \nmid w_i.
\end{equation}

\paragraph{Chaos}%
\label{para:chaos}
The behavior of dynamical systems can range from ordered to chaotic. The framework of dynamical systems lacks a precise and universal definition of chaos. However, there is
widespread agreement that chaotic behavior is based on \textit{sensitivity}, \textit{transitivity} and \textit{regularity}~\cite{Devaney2020}.
Manzini and Margara~\cite{Manzini1999a} identified five classes of increasing degree of chaos for linear \glspl{ca}: \textit{equicontinuous} \glspl{ca}, \textit{sensitive} but
not \textit{transitive} \glspl{ca}, \textit{transitive} but not \textit{strongly transitive} \glspl{ca}, \textit{strongly transitive} but not \textit{positively expansive} \glspl{ca}, and \textit{positively expansive} \glspl{ca}.
Since for linear \glspl{ca}, \textit{transitivity} implies \textit{sensitivity} and \textit{surjectivity}, whereby the latter is in turn equivalent to \textit{regularity},
\textit{transitive} linear \glspl{ca} can be classified as topologically chaotic~\cite{Cattaneo1997a}.

\subsubsection{Error Metric}
To be able to compare different models, we use the \gls{mse}
\begin{equation}
    \mse(\vec{y}, \bar{\vec{y}}) = \frac{1}{n} \sum^{n}_{i=1} {\left( \bar{y}_i - y_i\right)}^2
\end{equation}
and the \gls{nmse}
\begin{equation}
    \nmse(\vec{y}, \bar{\vec{y}}) = \frac{\mse(\vec{y}, \bar{\vec{y}})}{\Var(\bar{\vec{y}})}
\end{equation}
with the ground truth \({\bar{\vec{y}}}\) and the prediction of the model \({\vec{y}}\).

\section{Refined ReCA Architecture}%
\label{sec:reca_proposed}

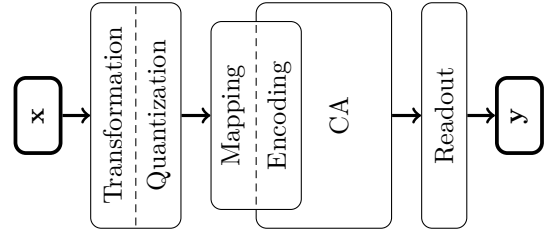
\begin{figure}[!t]
    \centering
    {
    \tikzstyle{inout} = [rectangle, draw, fill=white, rounded corners, minimum height=1cm, minimum width=0.6cm, line width=0.5mm]
    \tikzstyle{layer-split} = [rectangle, draw, fill=white, rounded corners, minimum height=2.5cm, minimum width=1.2cm]
    \tikzstyle{layer-split-large} = [rectangle, draw, fill=white, rounded corners, minimum height=3cm, minimum width=1.2cm]
    \tikzstyle{layer} = [rectangle, draw, fill=white, rounded corners, minimum height=3cm, minimum width=0.6cm]
    \tikzstyle{ca} = [rectangle, draw, fill=white, rounded corners, minimum height=3cm, minimum width=1.8cm]

    \begin{tikzpicture} [node distance = 1cm, auto]
        \node[inout] (in) {};
        \node[align=center,rotate=90] at (in.center) {\(\vec{x}\)};

        \node[layer-split-large, right of=in, xshift=3mm] (trans) {};
        \node[align=center,rotate=90,yshift=-3mm] at (trans.west) {Transformation};
        \node[align=center,rotate=90,yshift=3mm] at (trans.east) {Quantization};
        \draw[densely dashed, shorten <=1.5, shorten >=1] (trans.north) -- (trans.south);

        \node[ca, right of=trans, xshift=15mm] (ca) {};
        \node[align=center,rotate=90, yshift=-2mm] at (ca.center) {CA};

        \node[layer-split, right of=trans, xshift=6mm] (split) {};
        \draw[densely dashed, shorten <=1.5, shorten >=1] (split.north) -- (split.south);

        \node[align=center,rotate=90,yshift=-3mm] at (split.west) {Mapping};
        \node[align=center,rotate=90,yshift=3mm] at (split.east) {Encoding};

        \node[layer, right of=ca,xshift=6mm] (read) {};
        \node[align=center,rotate=90] at (read.center) {Readout};

        \node[inout, right of=read] (out) {};
        \node[align=center,rotate=90] at (out.center) {\(\vec{y}\)};

        \draw [->, very thick] (in) -- (trans);
        \draw [->, very thick] (trans) -- (split);
        \draw [->, very thick] (ca) -- (read);
        \draw [->, very thick] (read) -- (out);
    \end{tikzpicture}
}
    \caption{Refined \gls{reca} architecture}%
    \label{fig:reca_proposed}
\end{figure}

Based on the \gls{reca} architecture initially published by Yilmaz~\cite{Yilmaz2014}, we refine our view on the architecture to be able to more precisely define the different
computational steps within the \gls{reca} model.
In the rest of this paper, without loss of generality, we only consider the case of one-dimensional time series \({ \vec{x} = (x^{(0)},\ldots,x^{(T-1)}) }\) with \({ x^{(t)} \in [-1, 1] }\).
If n-dimensional data should be used, the transformation, quantization, mapping, and encoding layers are adjusted to the input dimension. For data \({ x^{(t)} \notin [-1, 1] }\), the transformation and quantization needs to be adopted.

We split the encoding layer (\cref{fig:reca_original}) into different parts since it fulfills several different and independent tasks.
The refined \gls{reca} architecture is depicted in \cref{fig:reca_proposed}.
The input data is fed into the transformation layer (\cref{subsec:transformation}), which prepares the data for the following quantization.
The transformation layer can also be used to run any transformation functions, e.g., tangens hyperbolicus, on the input data.
After the input is transformed, we need to quantize it to the allowed states \({x_q \in \intnum_m}\).
This is done by the quantization layer (\cref{subsec:quant}). Note that the transformation and quantization layers often work together to achieve the desired \({x_q}\).
The quantized input \({x_q}\) is then passed to the mapping layer (\cref{subsec:mapping}) and then the encoding layer (\cref{subsec:enc}).
The mapping layer selects the \gls{ca} cells into which the quantized input should be encoded. The encoding layer then executes the encoding.
After the \gls{ca} in the reservoir updated the cells for a fixed number of iterations, the states of the \gls{ca} are used by the readout layer (\cref{subsec:readout}) to calculate the \gls{reca} model output \({\vec{y}^{(t)}}\).

In \cref{subsec:reca_computations}, we will combine the aforementioned layers to the \gls{reca} model and describe a complete iteration of the model for an inference time step.

To improve the readability of the definitions, the superscript \({(t)}\) is removed in the rest of this section when the time context is clear.

\subsection{Transformation}%
\label{subsec:transformation}
We separate the transformation layer into two steps. First, we apply a transformation function \({\tilde{\vec{x}}_{\tau}=\tau(\vec{x})}\) to the input. Second, we scale the transformed
input to the range \({ x_{\tau} \in [0, m-1] }\) since we require this input range in the subsequent quantization layer.
For our setup, we analyzed the following transformation methods:
\begin{itemize}
    \item \textit{complement}
        \begin{equation}
            \tilde{x}_{\tau} =
            \begin{cases}
                x,&{\text{if}\ x \in [0,1]}, \\
                2+x,&{\text{otherwise.}}
            \end{cases}
        \end{equation}

    \item \textit{gray} and \textit{scale\_offset}
        \begin{equation}
            \tilde{x}_{\tau} = x + 1
        \end{equation}

    \item \textit{sign\_value}
        \begin{equation}
            \tilde{x}_{\tau} =
            \begin{cases}
                x,&{\text{if}\ x \in [0,1]}, \\
                -x+1,&{\text{otherwise.}}
            \end{cases}
        \end{equation}
\end{itemize}

Rescaling is then done using
\begin{equation}
    x_{\tau} = \frac{m-1}{2} \tilde{x}_{\tau}.
\end{equation}

The idea of the different transformation approaches is to mimic different floating-point to fixed-point conversion methods.
Using the \textit{complement} transformation will represent the numbers similar to a two's complement while \textit{sign\_value} uses a binary sign and value representation.
The \textit{scale\_offset} approach will shift the input range to only positive numbers and then use the default binary representation.
The \textit{gray} transformation uses the same shift but will encode the values using gray code.
The conversion to gray code is only correct if m is a power-of-two, otherwise neighboring values might not differ in only one bit.
\subsection{Quantization}%
\label{subsec:quant}
To quantize the input values, we use the typical rounding approach
\begin{equation}
    \tilde{x}_q =
    \begin{cases}
        0,&{\text{if}\ x_{\tau} \in [0,0.5)}, \\
        1,&{\text{if}\ x_{\tau} \in [0.5,1.5)}, \\
        2,&{\text{if}\ x_{\tau} \in [1.5,2.5)}, \\
        &\vdots \\
        m-1,&{\text{if}\ x_{\tau} \in [m-1.5,m-1]}.
    \end{cases}
    \label{eq:quantization}
\end{equation}

In case of \textit{gray} transformation, the quantized input \({\tilde{x}_q}\) is transformed once more, leading to the final quantized value\footnote{This could also be achieved by changing the quantization function.}
\begin{equation}
    x_q =
    \begin{cases}
        \tilde{x}_q \oplus (\tilde{x}_q >> 1) \mod m,&{\text{if \textit{gray}}},\\
        \tilde{x}_q,&{\text{else}}
    \end{cases}
    \label{eq:quantization2}
\end{equation}
with \({\oplus}\) representing the binary bitwise \textit{exclusive-or} and \({>>}\) is the binary right-shift operation.
The mod \({m}\) operation is only needed if m is not a power of \({2}\).

\subsection{Mapping}%
\label{subsec:mapping}
Yilmaz~\cite{Yilmaz2014} mentions that multiple random projections of the input into the reservoir are necessary to achieve low errors. However, instead of implementing multiple separate \gls{ca} reservoirs as in~\cite{Yilmaz2014}, we follow the design as described in~\cite{Babson2019} and subdivide a single \gls{ca} lattice into multiple parts.
Therefore, we divide the lattice of the \gls{ca} in the reservoir into \({N_r}\) compartments. Each compartment has the same number of \({N_c}\) cells. For example, a lattice of size
\({N=512}\) divided into \({N_r=16}\) compartments with \({N_c=32}\) cells each is described by the tuple \({ (N_r, N_c)=(16, 32) }\) with \({ N = N_r N_c }\).
The mapping layer selects the cells of the \gls{ca} that should receive the input value.
One cell is randomly selected out of each compartment, into which the input is encoded in the next step. This random mapping is fixed once and does not change. It can be modeled
as a vector \({\vec{p} \in \intmodring{m}^N}\) with the entries representing the cells of the \gls{ca} that shall receive the input set to \({x_q}\), and all other entries set to
zero (see \cref{fig:reca_computations} part \Romannum{1}).
\subsection{Encoding}%
\label{subsec:enc}
Since the mapping layer only defines into which cells the quantized input \({x_q}\) should be encoded, we have to define how the encoding is actually done.
For this, we use the following commonly used encoding functions.
Let \({\bar{s}_0}\) be the initial state of an individual cell in the reservoir that has been selected to receive \({x_q}\) via random mapping (see \cref{subsec:mapping}).
Then, the encoded cell state \({s_0}\) is defined by:

\begin{itemize}
    \item \textit{replacement} encoding~\cite{Yilmaz2014}
        \begin{equation}
            s_0 = x_q
        \end{equation}

    \item bitwise \textit{xor} encoding~\cite{Nichele2017,Margem2020}
        \begin{equation}
            s_0 = x_q \oplus \bar{s}_0 \mod m.
            \label{eq:enc_xor}
        \end{equation}
\end{itemize}
Note that if \({m}\) is a power of two, then the mod \({m}\) operation in \cref{eq:enc_xor} can be omitted.

Additionally, we analyzed the following new encoding functions:
\begin{itemize}
    \item \textit{additive} encoding
        \begin{equation}
            s_0 = x_q + \bar{s}_0 \mod m
        \end{equation}

    \item \textit{subtractive} encoding
        \begin{equation}
            s_0 = \left|x_q-\bar{s}_0\right|
        \end{equation}
\end{itemize}

The states of the cells not selected by the mapping layer do not change during the encoding process.
The \textit{replacement} encoding overwrites the information stored in the affected cells of the \gls{ca} with the new input.
This is different for the \textit{xor} encoding, which combines the new input with the current cell states and is, next to \textit{replacement} encoding, commonly used in \gls{reca}.
In order to analyze the influence of small changes in the encoding, we use the \textit{additive} and \textit{subtractive} encoding schemes, which slightly differ from the \textit{xor} encoding.
\subsection{Readout}%
\label{subsec:readout}
The readout layer is typically the weighted sum of the reservoir output
\begin{equation}
    \vec{y} = \Mat{U}\vec{r}+\vec{b}
\end{equation}
with the weight matrix \({\Mat{U}}\) and bias \({\vec{b}}\). The reservoir output \({\vec{r}}\) can be a single state vector \({\vec{s}}\) of the \gls{ca},
but is usually chosen to be a concatenation of the \gls{ca} state at multiple iteration steps~\cite{Yilmaz2014}.
Since \({\Mat{U}}\) and \({\vec{b}}\) are the only trainable parameter in the \gls{reca} model, a simple linear regression can be used.
To simplify the notation, it will be assumed that the input to the readout layer \({\vec{r}}\) has a
\({1}\) appended to also include the bias \({\vec{b}}\) in the weight matrix \({\Mat{U}}\).

To train the \gls{reca} model, the reservoir output \({\vec{r}^{(t)}}\) is concatenated for each input \({\vec{x}^{(t)}}\) into \({\Mat{R}}\). Furthermore, the ground truth solutions \({\bar{\vec{y}}^{(t)}}\) are concatenated in the same way to generate \({\bar{\Mat{Y}}}\).
When using ordinary least squares, the weight matrix \({\Mat{U}}\) can be calculated by
\begin{equation}
    \Mat{U} = {\left(\Mat{R}^T\Mat{R}\right)}^{-1}\Mat{R}^T \bar{\Mat{Y}}.
\end{equation}
There are many different adoptions to the linear regression algorithm. For example, Tikhonov regularization~\cite{Golub1999},
also called L2 regularization, can be added, resulting in Ridge Regression~\cite{Hoerl1970}.
It is also possible to run linear regression in an online and sequential approach~\cite{Liang2006}.
We use Ridge Regression in our experiments.
\subsection{ReCA Computations}%
\label{subsec:reca_computations}

\begin{figure}[!t]
    \centering
    {
    \input{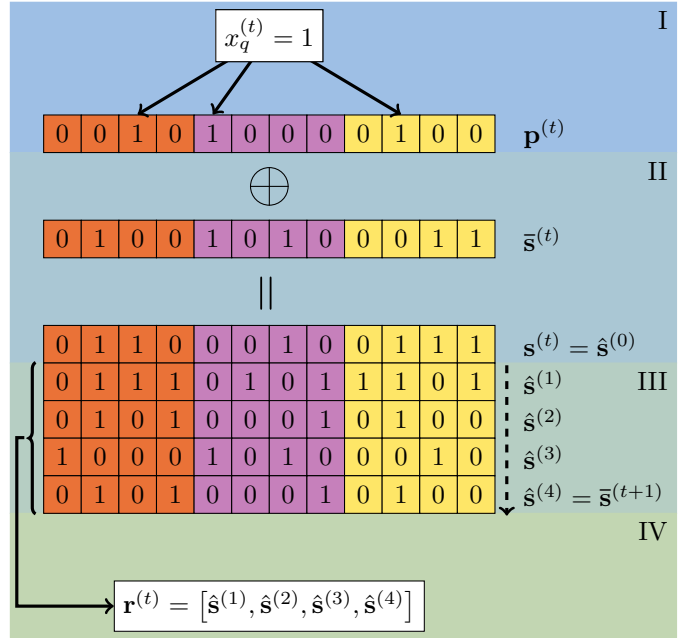}
    \def\compartmentcolors{tum-red,tum-pink-1,tum-yellow-2}
    \tikzstyle{xor} = [circle, draw, minimum width=\cellsize, minimum height=\cellsize]

    \newcommand{\recacell}[4]{
        \setsepchar{,}
        \readlist\coordlist#2
        \begin{scope}[auto]
            \node[cell, fill=#3, anchor=north] (#1) at (\coordlist[1],\coordlist[2]) {#4};
        \end{scope}
    }
    \newcommand{\compartment}[5]{
        \begin{scope}[auto]
            \setsepchar{,}
            \readlist\captionlist#5
            \readlist\coordlist#2
            \tikzmath{
                int \i;
                for \i in {1,...,#3}{
                    \ia = \i-1;
                    coordinate \cellpos;
                    \cellpos = (\coordlist[1], \coordlist[2]) + (\ia * \cellsize, 0);
                    \cellcaption = "\captionlist[\i]";
                    \celllabel = "#1cell\i";
                    {\recacell{\celllabel}{{\cellposx,\cellposy}}{#4}{\cellcaption}};
                };
            }
        \end{scope}
    }
    \newcommand{\recaconfig}[7]{
        \begin{scope}[auto]
            \setsepchar{,}
            \readlist\coordlist#2
            \readlist\colorlist#5
            \setsepchar{\\/,}
            \readlist\cellcaptions#6
            \tikzmath{
                int \i;
                for \i in {1,...,#3}{
                    \ia = \i-1;
                    coordinate \compos;
                    \compos = (\coordlist[1], \coordlist[2]) + (\ia * #4 * \cellsize, 0);
                    \comcolor = "\colorlist[\i]";
                    \comcaptions = "\cellcaptions[\i]";
                    \compartmentlabel = "#1com\i";
                    {
                        \compartment{\compartmentlabel}{{\composx, \composy}}{#4}{\comcolor}{\comcaptions}
                    };
                };
                {\node[align=center,anchor=west] (#1caption) at ($(#1com#3cell#4.east) + (0.25, 0)$) {#7};};
            }
        \end{scope}
    }
    \pgfdeclarelayer{background}
    \pgfdeclarelayer{foreground}
    \pgfsetlayers{background, main, foreground}
    \begin{tikzpicture}[node distance=0, auto]
        \begin{pgfonlayer}{background}
            \node[rectangle, fill=gradient-blue-1!60, minimum width=\linewidth, minimum height=2.0cm, anchor=north west, draw=gradient-blue-1!60]  (part1) at (-0.7,0.1) {};
            \node[anchor=north east] (part1label) at (part1.north east) {\Romannum{1}};
            \node[rectangle, fill=gradient-blue-2!60, minimum width=\linewidth, minimum height=2.80cm, anchor=north west, draw=gradient-blue-2!60] (part2) at (-0.7,-1.9) {};
            \node[anchor=north east] (part2label) at (part2.north east) {\Romannum{2}};
            \node[rectangle, fill=gradient-blue-3!60, minimum width=\linewidth, minimum height=2.0cm, anchor=north west, draw=gradient-blue-3!60] (part3) at (-0.7, -4.7) {};
            \node[anchor=north east] (part3label) at (part3.north east) {\Romannum{3}};
            \node[rectangle, fill=gradient-blue-4!60, minimum width=\linewidth, minimum height=1.65cm, anchor=north west, draw=gradient-blue-4!60] (part4) at (-0.7, -6.7) {};
            \node[anchor=north east] (part4label) at (part4.north east) {\Romannum{4}};
        \end{pgfonlayer}
        \begin{pgfonlayer}{foreground}
            \tikzmath{
                \totalwidth = (\ncompartments * \ncells);
                \halfwidth = (\ncompartments * \ncells - 1)/2;
                coordinate \chalf, \cxt, \cres, \cxor, \ceq;
                \chalf = (0,0) + (\halfwidth*\cellsize,0);
                \cxt = (\chalfx,\chalfy)+(0,0);
                \cxor = (\chalfx,\chalfy) - (0, 2.1);
                \ceq = (\chalfx, \chalfy) - (0, 3.4);
                \cres = (\chalfx, \chalfy) - (0, 7.6);
                {
                    \node[rectangle,draw,align=center, anchor=north, fill=white] (xt) at (\chalf) {\(x_q^{(t)}=1\)};
                    \recaconfig{con1}{{0,-1.4}}{3}{4}{{\compartmentcolors}}{{0,0,1,0\\1,0,0,0\\0,1,0,0}}{\(\vec{p}^{(t)}\)}
                    \node[xor, anchor=north] (xor) at (\cxor) {};
                    \draw[] (xor.north) -- (xor.south);
                    \draw[] (xor.west) -- (xor.east);
                    \recaconfig{con2}{{0,-2.8}}{\ncompartments}{\ncells}{{\compartmentcolors}}{{0,1,0,0\\1,0,1,0\\0,0,1,1}}{\(\vec{\bar{s}}^{(t)}\)}
                    \node[align=center,font=\huge,rotate=90, anchor=east] (eq) at (\ceq) {\(=\)};
                    \recaconfig{con3}{{0,-4.2}}{\ncompartments}{\ncells}{{\compartmentcolors}}{{0,1,1,0\\0,0,1,0\\0,1,1,1}}{\(\vec{s}^{(t)} = \hat{\vec{s}}^{(0)}\)}
                    \recaconfig{con4}{{0,-4.7}}{3}{4}{{\compartmentcolors}}{{0,1,1,1\\0,1,0,1\\1,1,0,1}}{\(\hat{\vec{s}}^{(1)}\)}
                    \recaconfig{con5}{{0,-5.2}}{3}{4}{{\compartmentcolors}}{{0,1,0,1\\0,0,0,1\\0,1,0,0}}{\(\hat{\vec{s}}^{(2)}\)}
                    \recaconfig{con6}{{0,-5.7}}{3}{4}{{\compartmentcolors}}{{1,0,0,0\\1,0,1,0\\0,0,1,0}}{\(\hat{\vec{s}}^{(3)}\)}
                    \recaconfig{con7}{{0,-6.2}}{3}{4}{{\compartmentcolors}}{{0,1,0,1\\0,0,0,1\\0,1,0,0}}{\(\hat{\vec{s}}^{(4)} = \vec{\bar{s}}^{(t+1)}\)}
                    \draw[->, very thick] (xt.center) -- (con1com1cell3.north);
                    \draw[->, very thick] (xt.center) -- (con1com2cell1.north);
                    \draw[->, very thick] (xt.center) -- (con1com3cell2.north);
                    \node[rectangle,draw,align=center, anchor=north, fill=white] (xt) at (\chalf) {\(x_q^{(t)}=1\)};
                    \node[rectangle,draw,align=center,anchor=north, fill=white] (res) at (\cres) {\(\vec{r}^{(t)} = \left[ \hat{\vec{s}}^{(1)}, \hat{\vec{s}}^{(2)}, \hat{\vec{s}}^{(3)}, \hat{\vec{s}}^{(4)} \right]\)};
                };
            }
            \draw[->, very thick,dashed] ($(con3caption.south west) - (0.1, 0)$) to ($(con7caption.south west) - (0.1, 0)$);
            \draw[decorate,decoration={brace,mirror},very thick] ($(con4com1cell1.north west) - (0.1cm,0)$) to ($(con7com1cell1.south west) - (0.1cm,0)$);
            \draw[->, very thick] ($(con4com1cell1.north west) - (0.25cm,2*\cellsize)$) -- ++(-0.1cm,0) |- (res.west);
        \end{pgfonlayer}
    \end{tikzpicture}
}
    \vspace*{-1em}
    \caption{Example of \gls{reca} computation for an input sample \({ x_q^{(t)}=1 }\) with an elementary rule 90 \gls{ca}, \({(N_r,N_c)=(3,4)}\), \textit{xor} encoding
    and \({I=4}\) steps. The state of the \gls{ca} after the \({i\textsuperscript{th}}\) iteration is denoted by \({\hat{\vec{s}}^{(i)}}\). The colors of the lattice indicate the three compartments.}%
    \label{fig:reca_computations}
\end{figure}

For each input sample \({x^{(t)}}\), the \gls{reca} model performs several steps.
The whole process is shown for an elementary rule 90 \gls{ca}~\cite{Wolfram1983} and \({I=4}\) iterations in \cref{fig:reca_computations}.
First, the input is transformed and quantized according to \cref{subsec:transformation,subsec:quant} resulting in the quantized sample \({x_q^{(t)}}\).
Next, \({x_q^{(t)}}\) is mapped to the reservoir cells by the mapping vector \({\vec{p}^{(t)}}\) as described in \cref{subsec:mapping}.
As an example, \cref{fig:reca_computations} (part \Romannum{1}) shows the random mapping of a sample \({x_q^{(t)} = 1}\) on the elements of \({\vec{p}^{(t)}}\),
such that each compartment receives the input at one randomly selected cell.
After the mapping, the quantized input has to be encoded into the initial state
\({\vec{\bar{s}}^{(t)}}\) of the reservoir at time \({t}\), resulting in the encoded initial state \({\vec{s}^{(t)}}\) as described in \cref{subsec:enc}.
In \cref{fig:reca_computations} (part \Romannum{2}), this is depicted for the \textit{xor} encoding.
The encoded state \({\vec{s}^{(t)}}\) forms the initial state \({\hat{\vec{s}}^{(0)}}\) for the \gls{ca}, which can then be executed for a fixed number of iterations \({I \in \natnum^{+}}\),
such that \({\hat{\vec{s}}^{(0)}}\) evolves under the repeated application of the linear \gls{ca} rule to \({\hat{\vec{s}}^{(I)}}\) (see \cref{fig:reca_computations} part \Romannum{3}).
After the execution of the \gls{ca} finishes, the reservoir outputs the concatenated \gls{ca} states \({\vec{r}^{(t)} = \left[ \hat{\vec{s}}^{(0)},\ldots, \hat{\vec{s}}^{(I)} \right]}\) (see \cref{fig:reca_computations} part \Romannum{4})
as mentioned in \cref{subsec:readout}.
The last state \({\hat{\vec{s}}^{(I)}}\) will be used as the initial reservoir state \({\vec{\bar{s}}^{(t+1)}}\) for the next input sample \({x^{(t+1)}}\).
\subsection{Hyperparameters}\label{subsec:hyperparameters}
Since the trainable parameters in the readout layer can be optimized using simple linear optimization techniques, a crucial step in designing \gls{reca} models is the choice of hyperparameters. In our analysis, we focus on the following general hyperparameters:
\begin{itemize}
    \item Number of states \(m\) of the \gls{ca}: This has an influcence on the operation domain of the \gls{ca} since \({\intnum_{m}}\) is either a field (if \(m\) is prime) or a ring (if \(m\) is non-prime). It significantly affects the mathematical properties and thus the dynamic behavior of the \gls{ca}. Since \(m\) defines the number of possible states of each cell, it also influences the linear separability of the reservoir output in the readout layer.
    \item True neighborhood \(\hat{n}\): The size of the neighborhood influences the expansion rate of local information on the lattice and thus also affects the dynamic behavior.
    \item Lattice size \(N\): This impacts the size of the dynamical system and thus affects the complexity and of the \gls{ca}
    \item Subdivision of the lattice into \(N_r\) compartments with \(N_c\) cells each: This influences the mapping of the input samples onto the reservoir cells.
    \item Number of iterations \(I\) of the \gls{ca} per input sample: This influences the degree of interactions between the cells per input sample.
    \item Transformation and quantization: This choice of transformation and quantization functions define how the input data is presented to the dynamical system.
    \item Mapping and encoding: The mapping and encoding methods define how the input is inserted into the state of the dynamical system.
\end{itemize}

Next to the general hyperparameters, an increased importance receives the hyperparameter \(F\), i.e., the global rule of the \gls{ca}, because it essentially defines the fundamental basis of the dynamics and topological properties of the \gls{ca}. As the rule space of linear \glspl{ca} grows exponentially with respect to \(m\) and \(\hat{n}\), it is vital to receive guidance when it comes to hyperparameter selection in the design process of \gls{reca} models. Since we restrict or analysis to linear rules, we term the respective framework as \gls{relica}.

It is important to note that all of these hyperparameters have interdependent effects on the overall behaviour of the \gls{ca} and in turn on the performance of the \gls{relica} model in time series processing tasks.
\section{Proposed ReLiCA Design Algorithm}%
\label{sec:relicada}

Since guidance in the choice of hyperparameters would greatly speed-up and assist in the design of \gls{relica} models, we propose the \acrfull{relicada}. We start with a short analysis of the influence of the \gls{ca} rules and transformation, quantization, mapping, and encoding layers on the \gls{relica} model performance in \cref{subsec:rule_impact} before introducing our \gls{relicada} in \cref{subsec:relicada}.

\subsection{CA Impact on ReLiCA Model Performance}%
\label{subsec:rule_impact}

\begin{figure}[!t]
    \centering
    \input{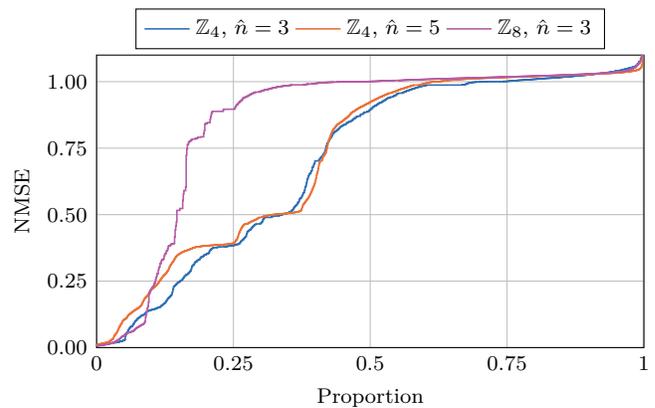}
    \caption{Empirical cumulative distribution functions for \textit{MG\_25}. The different configurations have the following number of data points: 960, 15360, 8064.}%
    \label{fig:ecdf}
\end{figure}

The choice of the \gls{ca} rule and the choice of transformation, quantization, mapping, and encoding functions significantly impact the overall \gls{relica} model performance.
We depict the \(\nmse\) for different \gls{relica} models for the \textit{MG\_25} dataset (see \cref{subsec:datasets}) in \cref{fig:ecdf} (other datasets produce similar results).
For this analysis, we ran all possible \gls{ca} rules with all combinations of the transformation and quantization configurations (\textit{complement}, \textit{gray}, \textit{scale\_offset}, \textit{sign\_value})
and the encoding functions (\textit{additive}, \textit{replacement}, \textit{subtractive}, \textit{xor}).
The empirical cumulative distribution function specifies the proportion of \gls{relica} models with the same or lower \(\nmse\).
As the figure shows, only a tiny percentage of all \gls{relica} models come close to the optimal performance for the chosen \(m\) and \(\hat{n}\) (lower left part of \cref{fig:ecdf}).
To the best of our knowledge, up to date, there are no clear rules or guidelines on how to select the linear \gls{ca}. Thus, obtaining a well-performing \gls{relica} model remained challenging.
Because of this, we developed an algorithm that pre-selects promising \gls{ca} rules. We will introduce this algorithm in the following subsection.
\subsection{ReLiCA Design Algorithm}%
\label{subsec:relicada}

We propose the \acrfull{relicada} to assist in the design of \gls{relica} models.
\gls{relicada} selects combinations of \gls{ca} rules, transformation, quantization, mapping, and encoding functions that will likely lead to well-performing \gls{relica} models.
The main idea is to limit the search space of linear \gls{ca} rules from \({m^n}\) (see \cref{subsec:ca}) to a small number of promising rules and to select matching
transformation, quantization, mapping, and encoding functions. Another purpose of \gls{relicada} is to be able to identify \gls{relica} models that produce low errors on a wide
range of different datasets, and not only on a single pathological dataset like the 5-bit memory task.

\gls{relicada} is based on the evaluation of thousands of train-test runs followed by a mathematical analysis of linear \gls{ca} properties.
Our approach was to exhaustively test the performance of \gls{relica} models with almost all combinations of the abovementioned transformation, quantization, mapping, and encoding methods over the complete rule search space of several linear \gls{ca} configurations (\({\hat{n}}\), \({m}\), \({N}\) and \({I}\)) on several different datasets.
As depicted in \cref{fig:ecdf}, only a tiny percentage of all \gls{relica} models achieve low errors, hindering random and heuristic search approaches, especially for complex \glspl{ca} (larger \(m\) or \(\hat{n}\)).
Our experiments indicate that specific conditions on the choice of the model's hyperparameters lead to an improvement in performance. For example, some transformation and quantization approaches are more robust against hyperparameter changes than others, and most of the generally good performing linear \gls{ca} rules share common mathematical properties (see
\cref{sec:results} for a detailed discussion of the results of the experiments). We identified these common rule properties and described them in terms of the mathematical parameters as defined in \cref{subsec:math_params}.
The result is \gls{relicada}, a set of selection rules, which are applied to the hyperparameters of \gls{relica} models. It limits the large number of all possible configurations to a small number of promising candidate models. A crucial part is the pre-selection of only very few linear \gls{ca} rules that are among the top-performing
rules in the overall rule space.
In doing so, \gls{relicada} enormously reduces the design time of \gls{relica} models because it prevents the need to undergo an exhaustive search over the whole linear \gls{ca} rule space, which is not feasible, especially for more complex \gls{ca}. Instead, \gls{relicada} enables the aimed testing of a few promising models that are sharply defined by the following conditions.

We use the definitions stated in \cref{subsec:math_params} to describe \gls{relicada}.
We limited our analysis to \({|\mathscr{P}| \leq 2}\), which will also be assumed in the description of the rule selection algorithm.
This was done since we are primarily interested in \({\intnum_m}\) with a single prime factor.
Some of the proposed rules might also work for the case \({|\mathscr{P}| > 2}\) or might be generalized, but no verification was done for that.

\subsubsection{ReLiCADA Design Rules}
The \gls{relicada} selects configurations only if all of the following conditions are fulfilled:
\begin{subequations}%
    \label{eq:alg-allways}
    \begin{gather}
        \text{transformation} = \textit{scale\_offset} \label{eq:alg-trans}\displaybreak[0]\\
        \text{quantization} = \textit{scale\_offset} \label{eq:alg-quant}\displaybreak[0]\\
        \text{mapping} = \textit{random} \label{eq:alg-map}\displaybreak[0]\\
        \text{encoding} = \textit{replacement} \label{eq:alg-enc}\displaybreak[0]\\
        (\forall p \in \primeset)(\exists _1 w_i) : p \nmid w_i \label{eq:alg-injective}\displaybreak[0]\\
        \nentropy = 1 \label{eq:alg-nentropy}\displaybreak[0]\\
        \text{remove mirrored rules} \label{eq:alg-rev}
    \end{gather}
\end{subequations}
The following selection rules will only be used based on the choice of \({m}\):
\begin{itemize}
    \item if \({m}\) is not prime, i.e., \({\intnum_m}\) forms a ring:
        \begin{subequations}%
            \label{eq:alg-ring}
            \begin{gather}
                \exists_2 i : w_i \neq 0 \label{eq:alg-zero-weights}\\
                \forall w_i: (w_i \notin \primeset_w) \lor (w_i \in \{1, m-1\}) \label{eq:alg-prime-group} \\
                \forall w_i: (w_i \notin \bar{\primeset}_w) \lor (|\mathscr{S}^+(w_i)|=4) \label{eq:alg-subgroup}
            \end{gather}
        \end{subequations}

    \item if \({m}\) is prime, i.e., \({\intnum_m}\) forms a field:
        \begin{equation}
            \forall w_i: (w_i=0) \lor (\mathscr{S}^{\times}(w_i)=\intnum_m {\setminus} 0) \label{eq:alg-field}
        \end{equation}

    \item if \({|\mathscr{P}| = 2}\):
        \begin{equation}
            (\exists w_i, w_j): (p_1 \nmid w_i) \land (p_2 \nmid w_j) \label{eq:alg-two-entropy}\\
        \end{equation}

\end{itemize}

Conditions \cref{eq:alg-allways} will always be used independent of the choice of \({m}\) and \({\hat{n}}\),
while conditions \cref{eq:alg-ring,eq:alg-two-entropy,eq:alg-field} are only used dependent on the choice of \({\intnum_m}\).
If \({\intnum_4}\) is chosen, it is impossible to fulfill rule \cref{eq:alg-subgroup}. Because of this, it will be ignored for the \({\intnum_4}\) case.

The selection \cref{eq:alg-rev} between the rule \({\vec{w}}\) and its mirrored rule \({\hat{\vec{w}}}\) is made using the following condition
\begin{equation}
    \label{eq:sel_mirrow}
    \sum_{i=-r}^{-1}w_i \leq \sum_{i=1}^{r}w_i,
\end{equation}
which evaluates to true for only one of the two rules if \({\vec{w}\neq\hat{\vec{w}}}\). If \({\vec{w}=\hat{\vec{w}}}\), this condition will always be fulfilled.
If the condition is true, we choose \({\vec{w}}\) and otherwise \({\hat{\vec{w}}}\).
The selection between \({\vec{w}}\) and \({\hat{\vec{w}}}\) is not optimized to increase the performance and is only used to further reduce the number of selected rules.
Because of this, also other selection methods between \({\vec{w}}\) and \({\hat{\vec{w}}}\) than \cref{eq:sel_mirrow} are possible.

For any given \({N, m}\) and \({\hat{n}}\), \cref{code:relicada-rule-selection} implements the process of rule selection (see \cref{sec:pseudocode}).

\subsubsection{Reasoning Behind Design Rules}
The conditions \cref{eq:alg-trans,eq:alg-quant,eq:alg-map,eq:alg-enc} fix the transformation, quantization, mapping, and encoding methods to an evidently well-performing
combination (see \cref{subsec:preliminary_results}). The conditions \cref{eq:alg-injective,eq:alg-nentropy,eq:alg-ring,eq:alg-field,eq:alg-two-entropy} belong to the rule selection for the linear \gls{ca} reservoir.
To limit the number of selected rules even further, we added condition \cref{eq:alg-rev}.

Our experiments showed that nearly all of the generally well-performing rules are \textit{injective}. Because of this, we included \cref{eq:alg-injective} (see \cref{eq:injective}).
With this condition, the \gls{ca} is not only \textit{injective} but also \textit{surjective} and \textit{regular} (see \cref{eq:regular}).
Moreover, the \glspl{ca} do not have a transient phase because of the \textit{injectivity}~\cite{Voorhees2012}.
The \textit{injectivity} of the \gls{ca} results in them not being \textit{strong transitive} (see \cref{eq:strong-transitivity}) and not being \textit{positive expansive} (see \cref{eq:positive-expansive}).

Furthermore, it was clear that \({\nentropy}\) has a significant impact on the \gls{reca} model performance.
Using \cref{eq:alg-nentropy} ensures that the \gls{ca} is \textit{sensitive} (see \cref{eq:sensitivity}) as well as \textit{transitive}, \textit{ergodic}, and \textit{expansive} (see \cref{eq:transitivity}).
While this would also be the case for other \({\nentropy}\) values, \({\nentropy=1}\) resulted, in most cases, in the best performance and has the advantage of the smallest possible neighborhood \({\hat{n} \geq 3}\), which reduces the complexity of hardware implementations%
\footnote{other \({\nentropy}\) values may require \({\hat{n} \geq 5}\)}. Condition \cref{eq:alg-nentropy} also implies that the \gls{ca} is not \textit{equicontinuous} (see \cref{eq:equicontinuity}).

Conditions \cref{eq:alg-ring,eq:alg-field,eq:alg-two-entropy} were chosen to improve the \gls{reca} model performance and reduce the overall amount of rules. These conditions were not chosen based on mathematical characteristics.

\subsubsection{Edge of Chaos}
While there is, to the best of our knowledge, no analysis of the \gls{eoc} done in the \gls{reca} framework, it is broadly discussed for \glspl{ca}~\cite{Langton1990,Packard1988,Mitchell1993,Devaney2020,Teuscher2022}.
The \gls{eoc} can be compared to the \gls{eols} in the \glspl{esn} framework~\cite{Verstraeten2010}.
Verstraeten et.al.~\cite{Verstraeten2010} analyzed the connection of the Lyapunov exponents of a specific \gls{esn} model to its memory and non-linear capabilities.
Through these analyses, it was shown that \glspl{ca} have the highest computational power at the \gls{eoc} and \gls{esn} models at the \gls{eols}.

Using the five groups of \gls{ca} rules with increasing degree of chaos, as defined by Manzini and Margara~\cite{Manzini1999a} (see \cref{para:chaos}), we can see that the
\Glspl{ca} selected by \gls{relicada} all belong to the third group, implying that they exhibit a ``medium'' amount of chaos.
By the definitions of Devaney and Knudsen they are chaotic, but not expansive chaotic~\cite{Cattaneo1999}.
Since the \gls{ca} rules selected by \gls{relicada} are among the best performing rules, we suppose, without any proof, that this might correlate to the edge of chaos.

As an example, for the configuration \({m=4}\) and \({\hat{n}=3}\), \gls{relicada} selects, among others, the rule with \({\vec{w}=(0,2,1)}\), which is depicted in \cref{fig:rule_impulse} and \cref{fig:rule_random}. The two iteration diagrams show that this linear \gls{ca}, on the one hand, has memorization capabilities by shifting the initial state to the left. This left shift can also be interpreted as a transmission of local information along the lattice. On the other hand, it shows interactions of neighboring cells during iteration. These properties (storage, transmission, and interaction) constitute computational capabilities in dynamical systems~\cite{Langton1990, Teuscher2022}.
Generally, all selected \gls{relicada} \gls{ca} rules show similar behavior.

\subsubsection{Number of Rules Selected by ReLiCADA}
\begin{table}[!t]
    \caption{Number of rules selected by ReLiCADA}%
    \label{tab:rules_selected}
    \centering
    \begin{tblr}{hlines, colspec=|c|c|c|,}
        \({\intnum_m}\) & rules selected & {total linear rules \\ \({\hat{n}=3}\)}\\
        \({m=3}\) & 1 & \phantom{000}24\\
        \({m=4}\) & 4 & \phantom{000}60\\
        \({m=5}\) & 2 & \phantom{00}120\\
        \({m=6}\) & 2 & \phantom{00}210\\
        \({m=8}\) & 8 & \phantom{00}504\\
        \({m=9}\) & 8 & \phantom{00}720\\
        \({m=12}\) & 8 & \phantom{0}1716\\
        \({m=16}\) & 8 & \phantom{0}4080\\
        \({m=32}\) & 8 & 32736\\
    \end{tblr}
\end{table}
In \cref{tab:rules_selected}, the number of rules selected by \gls{relicada} and the number of all linear rules are listed for different \({m}\) and \({\hat{n}=3}\).
It is worth pointing out that the number of selected rules by \gls{relicada} is independent of \({\hat{n}}\), whereas the number of total rules depends on the chosen neighborhood \({\hat{n}}\).
From \cref{tab:rules_selected}, it is easy to see that \gls{relicada} reduces the number of rules to analyze by several orders of magnitude.
Hence, when designing a \gls{reca} model for a specific application, one does not have to check all rules in the rule
space, but only the few rules that are pre-selected by \gls{relicada}.
\section{Experiments}%
\label{sec:experiments}

We will now introduce the experimental setup used to verify and validate the performance of \gls{relicada}.
The datasets used are introduced in \cref{subsec:datasets} and the models, to compare the performance of \gls{relicada} to, in \cref{subsec:compared_models}.

\subsection{Datasets}%
\label{subsec:datasets}

In order to test the performance of the different hyperparameter configurations of the \gls{relica} models, we use datasets that have already been used in several other papers to compare different time series models. The following datasets can thus be regarded as benchmark datasets.
These datasets might not need fast inference times, one of the main advantages of \gls{reca} models, but are suitable choices for broad comparability with other studies.
All datasets are defined over discrete time steps with \(t \in \natnum\).
We use \(x(t)\) to describe the input to the model, and \(y(t)\) represents the ground truth solution.
The \(x\) and \(y\) values are rescaled to \([-1,1]\). Unless otherwise noted, the task is to do a one-step-ahead prediction, i.e., \(y(t) = x(t+1)\), using the inputs up to \(x(t)\).
The abbreviations used to name the datasets throughout the paper are denoted by \textit{(name)}.

\subsubsection{H\'{e}non Map}
The H\'{e}non Map (\textit{H\'{e}non}) was introduced in~\cite{Henon1976} and is defined as
\begin{equation}
    y(t) = x(t+1) = 1 - 1.4{x(t)}^2+0.3x(t-1) \text{.}
\end{equation}

\subsubsection{Mackey-Glass}%
\label{subsubsec:mg}
The Mackey-Glass time series uses the nonlinear time-delay differential equation introduced by~\cite{Mackey1977}
\begin{equation}
    \frac{dx}{dt} = \beta\frac{x(t-\tau) }{1+{x(t-\tau) }^{n}}-\gamma x(t)
\end{equation}
with \(\beta=0.2\), \(\gamma=0.1\), \(\tau=17\), and \(n=10\).
The task is to predict \(y(t)=x(t+1)\) using \(x(t)\) (\textit{MG}).
Furthermore, we use the prediction task \(y(t)=x(t+25)\) using \(x(t)\) (\textit{MG\_25}).

\subsubsection{Multiple Superimposed Oscillator}
The Multiple Superimposed Oscillator (MSO) is defined as
\begin{equation}
    x(t)=\sum_{i=1}^{n}\sin(\varphi_i t) \text{, with } t \in \natnum\text{.}
\end{equation}
The MSO12 dataset uses \(\varphi_1=0.2\), \(\varphi_2=0.331\), \(\varphi_3=0.42\), \(\varphi_4=0.51\), \(\varphi_5=0.63\),
\(\varphi_6=0.74\), \(\varphi_7=0.85\), \(\varphi_8=0.97\), \(\varphi_9=1.08\), \(\varphi_{10}=1.19\), \(\varphi_{11}=1.27\), and \(\varphi_{12}=1.32\) as defined in~\cite{Esposito2019}. We use the prediction tasks \(y(t)=x(t+1)\) (\textit{MSO}) and \(y(t)=x(t+3)\) (\textit{MSO\_3}) with \(x(t)\) as input.

\subsubsection{Nonlinear Autoregressive-Moving Average}
The Nonlinear Autoregressive-Moving Average was first introduced in~\cite{Atiya2000} as a time series dataset. We use the 10th order (\textit{NARMA\_10})
\begin{equation}
    \begin{split}
        x(t+1) = &{~}0.3x(t) + 0.05x(t) \sum_{i=0}^{9}\left( x(t-i) \right)\\
        &+ 1.5u(t-9)u(t)+0.1 \text{,}\\
    \end{split}
\end{equation}
the 20th order (\textit{NARMA\_20})
\begin{equation}
    \begin{split}
        x(t+1) = &{~}\tanh[0.3x(t) + 0.05x(t)\sum_{i=0}^{19}\left(x(t-i)\right)\\
        &+ 1.5u(t-19)u(t) + 0.01] + 0.2 \text{,}
    \end{split}
\end{equation}
and the 30th order (\textit{NARMA\_30})
\begin{equation}
    \begin{split}
        x(t+1) = &{~}0.2x(t) + 0.004x(t) \sum_{i=0}^{29}\left( x(t-i) \right)\\
        &+ 1.5u(t-29)u(t) + 0.201
    \end{split}
\end{equation}
versions as defined in~\cite{Chen2013}. The input \(u(t)\) is generated by a uniform \gls{iid} random variable in the interval \([0, 0.5]\).
The task is to predict \(x(t)\) using \(u(t)\).

\subsubsection{Nonlinear Communication Channel}
This dataset emulates a nonlinear communication channel and was introduced in~\cite{Jaeger2004} as
\begin{equation}
    \begin{split}
        q(t) = &{~}0.08u(t+2) - 0.12u(t+1) + u(t) + 0.18u(t-1) \\
        &- 0.1u(t-2) + 0.09 u(t-3) - 0.05u(t-4) \\
        &+ 0.04u(t-5) + 0.03u(t-6) + 0.01u(t-7) \\
        x(t) = &{~}q(t) + 0.036{q(t)}^2 - 0.011{q(t)}^3 \text{.}
    \end{split}
\end{equation}
The channel input \(u\) is a random \gls{iid} sequence sampled from \(\{-3, -1, 1, 3\}\). The task is to predict \(x(t-2)\) using \(u(t)\) (\textit{NCC}).

\subsubsection{Pseudo Periodic Synthetic Time Series}
Introduced by UC Irvine~\cite{Dua2017}, the dataset can be generated using
\begin{equation}
    x(t) = \sum_{i=3}^{7} \frac{1}{2^i} \sin \left( 2 \pi \left( 2^{2 + i} + rand(2^i) \right) * \frac{t}{10000} \right)
\end{equation}
as defined in~\cite{Park1999} (\textit{PPST}).

\subsubsection{Predictive Modeling Problem}
First introduced by Xue et.\ al.~\cite{Xue2007}, the dataset can be generated using
\begin{equation}
    x(t) = \sin(t+\sin(t)) \text{,\qquad with } t \in \natnum
\end{equation}
(\textit{PMP}).
\subsection{Compared Models}%
\label{subsec:compared_models}
To have a reference for the \gls{relica} model performance values, several state-of-the-art models were used as a baseline.
These models and their hyperparameters are established in this section.
In the following description, a parameter optimized during hyperparameter optimization is denoted by a range, e.g., \({[a, b]}\).
The results for these models are listed in \cref{tab:baseline}.

\subsubsection{Neural Networks}
These models were created using \texttt{TensorFlow} 2.8.0~\cite{Abadi2015} with the default settings unless otherwise noted.
The models have an \texttt{Input} layer and use a \texttt{Dense} layer as output. The hidden layers were adopted to the used model.
We used \texttt{Adam}~\cite{Kingma2014} as the optimizer, and the learning rate was \({[10^{-10}, 1]}\).
As a loss function, \mse{} is used.

The Recurrent Neural Network~\cite{Rumelhart1986} (\textit{RNN}) uses a \texttt{SimpleRNN} layer with 64 units with dropout \({[0, 1]}\) and recurrent dropout \({[0, 1]}\).

The \texttt{GRU} layer was used for the Gated Recurrent Unit NN~\cite{Cho2014} with 32 units and dropout \({[0, 1]}\).

The Long Short Term Memory NN~\cite{Hochreiter1997} uses the \texttt{LSTM} layer with 32 units and dropout \({[0, 1]}\).

The Neural Network~\cite{Rosenblatt1958} \textit{NN} model uses \({[1,4]}\) \texttt{Dense} layers with \({[1,64]}\) neurons per layer as hidden layers. The inputs to this model are the last 20 values of \({x(t)}\), which results in the vector \({\Vec{x}=[x(t-19), x(t-18), \ldots, x(t)]}\).

\subsubsection{RC Models}
We used an \gls{esn}, \gls{scr}, and \gls{dlr} model. All models use the \texttt{Scikit-learn} 1.1.2~\cite{Pedregosa2011} \texttt{Ridge} optimizer with an alpha \({[10^{-10}, 1]}\).

The \textit{ESN} model~\cite{Jaeger2001} uses the Tensorflow Addons \texttt{ESN} cell implementation embedded into our code. We used 128 units with a connectivity of \qty{10}{\percent}. The other parameters are input scale \({[0, 10]}\), input offset \({[-10, 10]}\), spectral radius \({[0, 1]}\), and leaky integration rate \({[0, 1]}\).

We implemented the \textit{SCR} and \textit{DLR} models according to~\cite{Rodan2011}. Both use 256 units, a spectral radius of \({[0, 1]}\), input scale \({[0, 10]}\), and input offset \({[-10, 10]}\).

\subsubsection{ReCA Models}
We used our implementation of our modified \gls{reca} architecture together with the non-linear \gls{ca} rules found by a \gls{ga} in~\cite{Babson2019}.
The lattice has a size of \({(16, 32)}\), and the \gls{ca} performs four iterations per input sample.
All rules found by Babson et al.~\cite{Babson2019} were analyzed using all combinations of the transformation and quantization configurations (\textit{complement}, \textit{gray}, \textit{scale\_offset}, \textit{sign\_value})
and the encoding functions (\textit{additive}, \textit{replacement}, \textit{subtractive}, \textit{xor}).
A Ridge optimizer is used for training. We call this model \textit{Babson}.

The \gls{reca} models were trained using \({100}\) parallel models with \({1100}\) time samples each. For testing and validation, \({1100}\) data points are used.
For training, testing, and validation, the first \({100}\) data points are used as initial transient and are not utilized.

\subsubsection{Linear Model}
A simple linear regression model using \texttt{Scikit-learn} was also evaluated. Like the \textit{NN}, the linear model, denoted by \textit{Linear}, has the last 20 values of \({x(t)}\) as input.

\subsubsection{Hyperparameter Optimization}
The hyperparameter optimization framework \texttt{Optuna}~\cite{Akiba2019} is used to optimize the hyperparameters of the models. We use the \texttt{TPESampler} algorithm with 100 runs per model.
For models using epoch-based training, early stopping was used. It was configured to stop the training if the loss is not decreasing by at least \({10^{-5}}\) with a patience of three epochs.

\subsubsection{Complexity}
One of the main advantages of \gls{reca} models is their low computational complexity. To compare the complexities of the different types of models, we approximated their computational complexities of the inference step.
This analysis was optimized for implementations on \glspl{fpga} without the usage of specialized hardware like multiply-accumulate units. Nevertheless, it should be a good indication also for other types of implementations.
Assuming two numbers \({a}\), \({b}\) that are represented by \({a'}\), \({b'}\) bits, we define the following complexities: addition and subtraction have a complexity of
\({\min(a',b')}\), whereas multiplication and division have a complexity of \({a' \times b'}\). For the additions and subtractions, we assume that the hardware does not need to deal with the \glspl{msb} of the larger number since these are zero in the smaller number.
For multiplication and division, we assume a shift-and-add implementation.
To approximate the complexity of the \textit{tanh} function, we use the seventh-order Lambert's continued fraction~\cite{Wall1948} as an approximation. We assume the same complexity for the \textit{sigmoid} function.
The \textit{ReLU} function has a complexity of zero.

We assume that the input and output have 32 bits, and all models use 32 bits to represent their internal states.
For the \gls{reca} models, the \gls{ca} uses the required number of bits to represent \({\intnum_m}\), and the readout layer also uses 32 bits.

The number of units in the different baseline models was chosen to make the overall model complexity similar to the tested \gls{relica} models.
Because of this, the number of units was not optimized for model performance.

\section{Results}%
\label{sec:results}

The results of our experiments can be divided into two phases. In the first phase, as described in \cref{subsec:preliminary_results},
we identified well-working choices for some of the general hyperparameters of the \gls{relica} model
that were fixed for the large number of experiments and exhaustive \gls{ca} rule performance analysis in phase two,
which is described in \cref{subsec:main_results}.
Detailed results of the experiments can be found in \cref{sec:app_relicada}.

\subsection{General Hyperparameters}%
\label{subsec:preliminary_results}

\begin{figure}[!t]
    \centering
    \begin{tikzpicture}
\begin{axis}[%
xmin=0, xmax=1,
xlabel={Proportion},
ytick distance=0.25,
xtick distance=0.25,
ymin=0, ymax=1.1,
ylabel={mean NMSE},
grid=major,
axis background/.style={fill=white},
legend style={legend cell align=left, align=left, draw=black, font=\footnotesize, at={(0.5,1.02)},anchor=south},
every y tick label/.append style={/pgf/number format/.cd,fixed,precision=2,fixed zerofill},
scaled y ticks=false,
legend columns=3,
width=\figurewidth,
height=\figureheight,
font=\footnotesize,
xtick style={draw=none},
ytick style={draw=none},
]

\addplot [semithick, tum-blue-brand, const plot mark right]
table {%
0.0 0.16745267338612735
0.016666666666666666 0.16745267338612735
0.03333333333333333 0.16745267388048796
0.05 0.17853565881384642
0.06666666666666667 0.1785678035277566
0.08333333333333333 0.17903407201502708
0.09999999999999999 0.17914208445169408
0.11666666666666667 0.17916909453015054
0.13333333333333333 0.17922177157637903
0.15 0.18643254111388963
0.16666666666666666 0.1864325412585642
0.18333333333333332 0.18806172084407752
0.19999999999999998 0.19041653083645121
0.21666666666666667 0.19831332497106308
0.23333333333333334 0.20030384310979074
0.25 0.2053752390058591
0.26666666666666666 0.2082310236943942
0.2833333333333333 0.6924691625138071
0.3 0.6924691652292885
0.31666666666666665 0.6924691654177619
0.3333333333333333 0.7276083106768226
0.35 0.7276083107130907
0.36666666666666664 0.7276083112510856
0.3833333333333333 0.8773516087256935
0.39999999999999997 0.8778032174361323
0.4166666666666667 0.8924364303514647
0.43333333333333335 0.8967088939617482
0.45 0.9003599930885212
0.4666666666666667 0.9014881245974098
0.48333333333333334 0.9031235504432545
0.5 0.9052876056253942
0.5166666666666667 0.9373444041198512
0.5333333333333333 0.9373444145314895
0.55 0.9373444146478276
0.5666666666666668 0.9373444146478276
0.5833333333333334 0.9373444397456578
0.6000000000000001 0.937344440767838
0.6166666666666667 0.9493266822503568
0.6333333333333334 0.9503405359689212
0.65 0.9505778403221191
0.6666666666666667 0.9509845788657533
0.6833333333333333 0.9516365280875342
0.7000000000000001 0.9521778775547197
0.7166666666666667 0.9522588927875238
0.7333333333333334 0.9524594989097878
0.75 1.0144607459015764
0.7666666666666667 1.0147528851450185
0.7833333333333333 1.01499899598303
0.8 1.0152966640869705
0.8166666666666668 1.0156280284781802
0.8333333333333334 1.0157439883780535
0.8500000000000001 1.0157662587889664
0.8666666666666667 1.016272594496759
0.8833333333333334 1.0166953034645767
0.9 1.0168090737016682
0.9166666666666667 1.0168417422464515
0.9333333333333333 1.0174886204740485
0.9500000000000001 1.0175474896470211
0.9666666666666667 1.0183282371479192
0.9833333333333334 1.0191487039125056
1.0 1.0192001072567407
};
\addlegendentry{\((16,32)\)}
\addplot [semithick, tum-red, const plot mark right]
table {%
0.0 0.16618679856410035
0.016666666666666666 0.16618679856410035
0.03333333333333333 0.1661867996754061
0.05 0.17479894669544657
0.06666666666666667 0.17546729250949675
0.08333333333333333 0.17582648878675053
0.09999999999999999 0.17751388309458255
0.11666666666666667 0.17764421642593395
0.13333333333333333 0.18001765838210487
0.15 0.18041543893036757
0.16666666666666666 0.18429154745759807
0.18333333333333332 0.18429154822159413
0.19999999999999998 0.18670504701025145
0.21666666666666667 0.18744738618496115
0.23333333333333334 0.18920829133705375
0.25 0.19254667393178773
0.26666666666666666 0.19556027674171736
0.2833333333333333 0.6924691614926584
0.3 0.6924691639607344
0.31666666666666665 0.6924691644453168
0.3333333333333333 0.7276083096455473
0.35 0.7276083110020869
0.36666666666666664 0.7276083111847094
0.3833333333333333 0.8948787514158124
0.39999999999999997 0.895524251299048
0.4166666666666667 0.9042339580564333
0.43333333333333335 0.9073360480066299
0.45 0.9086672728579095
0.4666666666666667 0.9099080728720189
0.48333333333333334 0.9164850888740501
0.5 0.9188162767937826
0.5166666666666667 0.9373444041516587
0.5333333333333333 0.9373444145314895
0.55 0.9373444145314895
0.5666666666666668 0.9373444145314895
0.5833333333333334 0.937344440767838
0.6000000000000001 0.937344440767838
0.6166666666666667 0.9803374577468863
0.6333333333333334 0.98043160803193
0.65 0.981545542425116
0.6666666666666667 0.9815553531701494
0.6833333333333333 0.9817878706185481
0.7000000000000001 0.9819543960859314
0.7166666666666667 0.9822260142014014
0.7333333333333334 0.9839506333830743
0.75 1.012379478889358
0.7666666666666667 1.0134561282806387
0.7833333333333333 1.014152314572263
0.8 1.0145503445705126
0.8166666666666668 1.0145735923917476
0.8333333333333334 1.0148413943041397
0.8500000000000001 1.0152228939210506
0.8666666666666667 1.0152676836043941
0.8833333333333334 1.0154294697773911
0.9 1.0156835523939265
0.9166666666666667 1.0161913268583498
0.9333333333333333 1.016428674565795
0.9500000000000001 1.0165246002424446
0.9666666666666667 1.016754735664075
0.9833333333333334 1.0173918130366926
1.0 1.0182714754395727
};
\addlegendentry{\((16,33)\)}
\addplot [semithick, tum-pink, const plot mark right]
table {%
0.0 0.16014919947080763
0.016666666666666666 0.16014919947080763
0.03333333333333333 0.16014920100984548
0.05 0.17435923898647526
0.06666666666666667 0.175631981411709
0.08333333333333333 0.17592347152021476
0.09999999999999999 0.17597360486477448
0.11666666666666667 0.1780216230262918
0.13333333333333333 0.17836004578743683
0.15 0.178360047312398
0.16666666666666666 0.17929290006068752
0.18333333333333332 0.1807948509013525
0.19999999999999998 0.18182969008186242
0.21666666666666667 0.18528087629049736
0.23333333333333334 0.18931598045203105
0.25 0.1934979700342879
0.26666666666666666 0.19770277961045288
0.2833333333333333 0.6924691601723362
0.3 0.6924691624289369
0.31666666666666665 0.6924691635921886
0.3333333333333333 0.7276083127751408
0.35 0.7276083135703121
0.36666666666666664 0.727608314941018
0.3833333333333333 0.8498787464811031
0.39999999999999997 0.8508459551292437
0.4166666666666667 0.8644478125941716
0.43333333333333335 0.8652066385564813
0.45 0.8660121562559188
0.4666666666666667 0.8683075882918561
0.48333333333333334 0.8762979266769065
0.5 0.8823694095661615
0.5166666666666667 0.9373444041314372
0.5333333333333333 0.9373444113355769
0.55 0.9373444113355769
0.5666666666666668 0.9373444123346173
0.5833333333333334 0.937344436647716
0.6000000000000001 0.9373444367242952
0.6166666666666667 0.9467086480351491
0.6333333333333334 0.9482884001432693
0.65 0.9483936290929832
0.6666666666666667 0.9487771981283726
0.6833333333333333 0.9489527666053706
0.7000000000000001 0.9492477328482686
0.7166666666666667 0.9495635784827244
0.7333333333333334 0.9503314507485093
0.75 1.0143797618361359
0.7666666666666667 1.014817744961689
0.7833333333333333 1.0148586488419058
0.8 1.0148722258771634
0.8166666666666668 1.0149770142301275
0.8333333333333334 1.0155479575609367
0.8500000000000001 1.0155948026225463
0.8666666666666667 1.0158797394581842
0.8833333333333334 1.0160121507946793
0.9 1.0164558252935971
0.9166666666666667 1.0165146856959433
0.9333333333333333 1.0166742873202401
0.9500000000000001 1.0168689687260903
0.9666666666666667 1.0172123846817782
0.9833333333333334 1.0175522082957293
1.0 1.0194828654268444
};
\addlegendentry{\((17,31)\)}
\end{axis}
\end{tikzpicture}%
    \caption{Influence of the lattice size \({(N_r, N_c)}\) on \gls{relica} with \({\intnum_4}\), \({\hat{n}=3}\), \({I=4}\) scale\_offset, and replacement.}%
    \label{fig:lattice}
\end{figure}
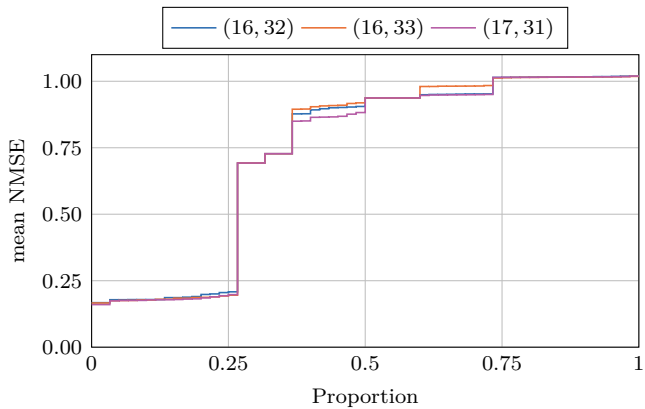

Since the main focus of our analysis lies in selecting suitable combinations of transformation, quantization, mapping, and encoding methods,
and linear \gls{ca} rules for the reservoir, we fixed some of the hyperparameters of the \gls{relica} framework to reduce the parameter space.
Therefore, we first analyzed the influence of the reservoir size  (\({N}\)) and the number of \gls{ca} iterations on the overall \gls{relica} performance.
The datasets used for the following analysis are MG, MG\_25, MSO, MSO\_3, NARMA\_10, NARMA\_20, NARMA\_30, NCC, PPST, PPST\_10, PMP (see \cref{subsec:datasets}).

To test the influence of the reservoir size, we tested the following lattice sizes: \({(16,32)}\), \({(16,33)}\), and \({(17,31)}\).
These were chosen since the total number of cells is similar, but their prime factor decomposition differs significantly.
The results for the \gls{relica} model using \textit{scale\_offset}, \textit{replacement}, \({\intnum_4}\), \({\hat{n}=3}\) and \({I=4}\) are depicted in \cref{fig:lattice}.
Other \gls{relica} models showed similar behavior. Since none of the lattice sizes is superior to the others, we used \({(16,32)}\) in the following experiments.
This was done since, for most hardware implementations, a power-of-two number of cells would most likely be suitable.

\begin{figure}[!t]
    \centering
    \input{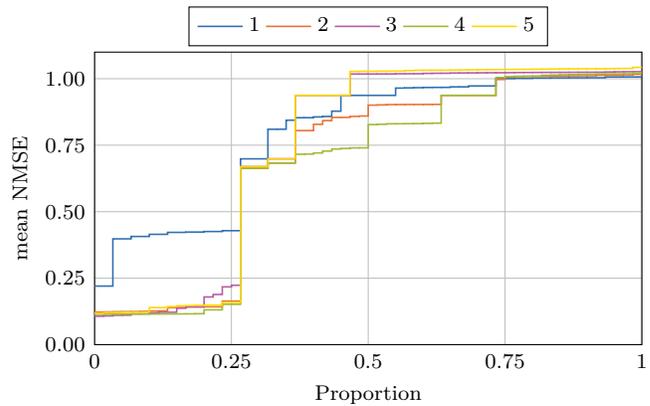}
    \caption{Influnece of the number of iterations \(I\) on \gls{relica} with \({\intnum_4}\), \({\hat{n}=3}\), (16,32), scale\_offset, and replacement.}%
    \label{fig:steps}
\end{figure}

To see the influence of the \gls{ca} iterations we tested the \gls{relica} model using \textit{scale\_offset}, \textit{replacement}, \({\intnum_4}\), \({\hat{n}=3}\), and \({(N_r, N_c) = (16,32)}\).
The results are shown in \cref{fig:steps}, and other configurations resulted in similar results.
Increasing the number of \gls{ca} iterations to \({>2}\) steps did not lead to a significant monotonic decrease in the overall \nmse{}. This is in line with the results by
Babson et al.~\cite{Babson2019}, where they achieved a success rate of \({\qty{99}{\percent}}\) in the 5-bit memory task for complex CA reservoirs and four iterations. In their
study, elementary (\({m=2}\), \({\hat{n}=3}\)) \glspl{ca} were found to require eight iterations.
However, Nichele et al.~\cite{Nichele2017a} show that several single elementary \gls{ca} rules also achieve a success rate of \({\qty{\ge 95}{\percent}}\) in the 5-bit memory task with only four iterations.
Since higher numbers of \gls{ca} iterations imply a higher computational complexity and longer training and testing times,
we fixed the number of iterations to four in all subsequent experiments.

\begin{figure}[!t]
    \centering
    \input{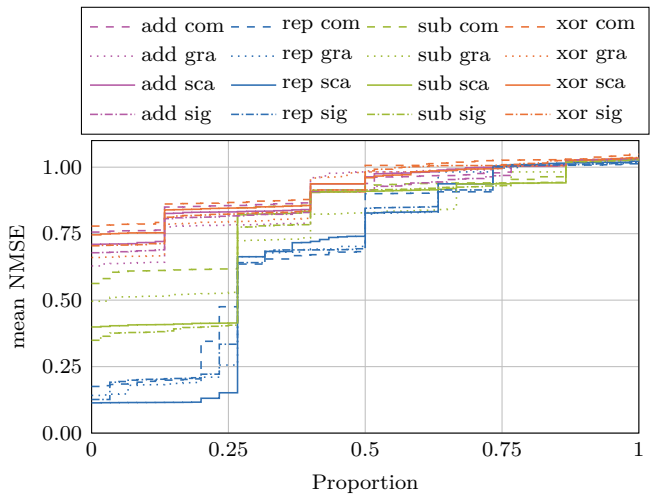}
    \caption{Comparison of the different transformation, quantization, mapping, and encoding functions using \gls{relica} with \({\intnum_4}\), \({\hat{n}=3}\) \({I=4}\), \({(N_r, N_c) = (16,32)}\).
    Used abbreviations: \textit{add}itive, \textit{rep}lacement, \textit{sub}tractive, \textit{xor}; \textit{com}plement, \textit{gra}y, \textit{sca}le\_offset, \textit{sig}n\_value.}%
    \label{fig:algs}
\end{figure}

Another finding is that the \textit{replacement} encoding together with the \textit{scale\_offset} transformation achieves low errors in most configurations and is thus the most stable encoding with respect to changing values of
the other hyperparameters. This can be seen in \cref{fig:algs}. Therefore, we fixed the transformation method to \textit{scale\_offset} and the encoding to random \textit{replacement}.

The random mapping generator's seeds, the only random element in the \gls{relica} model, were fixed to ensure reproducible results.
\subsection{Rule Selection}%
\label{subsec:main_results}

To analyze the performance of the \acrlong{relicada}, we used the following time-series benchmark datasets: MG, MG\_25, MSO, MSO\_3, NARMA\_10, NARMA\_20, NARMA\_30, NCC, PPST, PPST\_10, PMP (see \cref{subsec:datasets}).
To train the \gls{relica} models, a Ridge optimizer is used with \({\alpha=1}\), the default value for Scikit-learn. The number of states \(m\), neighborhood \(\hat{n}\), and local rule of the \gls{ca} are varied throughout the experiments.
We denote the models designed using \gls{relicada} by \gls{relica}* and the general class of \gls{relica} models using the whole set of possible linear \gls{ca} rules by \gls{relica}.
All individual performance values are listed in the \cref{sec:app_baseline,sec:app_relicada}.
We used a train-test split for the datasets to conduct our experiments. Unless otherwise noted, the test performance values are used.

\begin{figure}[!t]
    \centering
    \begin{tikzpicture}
\begin{axis}[
axis background/.style={fill=white},
xmin=-0.5, xmax=3.5,
xtick={0,1,2,3},
ytick distance=0.025,
xticklabels={{\(\intnum_4\), \(\hat{n}=3\)},{\(\intnum_4\), \(\hat{n}=5\)},{\(\intnum_8\), \(\hat{n}=3\)},{\(\intnum_{16}\), \(\hat{n}=3\)}},
ylabel={mean \(\nmse\)},
ymajorgrids,
ymin=0, ymax=0.125,
width=\figurewidth,
height=\figureheight,
font=\footnotesize,
xtick style={draw=none},
ytick style={draw=none},
every y tick label/.append style={/pgf/number format/.cd,fixed,precision=3,fixed zerofill},
legend style={legend cell align=left, align=left, draw=black, font=\footnotesize, at={(0.5,1.02)},anchor=south},
legend columns=2,
]
\fill[tum-blue-brand] (axis cs:-0.4,0) rectangle (axis cs:-0.133333333333333,0.111106682950032);
\fill[tum-blue-brand] (axis cs:0.6,0) rectangle (axis cs:0.866666666666667,0.109714501183067);
\fill[tum-blue-brand] (axis cs:1.6,0) rectangle (axis cs:1.86666666666667,0.0562844882453799);
\fill[tum-blue-brand] (axis cs:2.6,0) rectangle (axis cs:2.86666666666667,0.0455750941335713);
\addlegendimage{ybar,ybar legend,draw=tum-blue-brand,fill=tum-blue-brand}
\addlegendentry{ReLiCA\phantom{*} best}
\addlegendimage{ybar,ybar legend,draw=white,fill=white}
\addlegendentry{{}}
\fill[tum-red] (axis cs:-0.133333333333333,0) rectangle (axis cs:0.133333333333333,0.113124678499637);
\fill[tum-red] (axis cs:0.866666666666667,0) rectangle (axis cs:1.13333333333333,0.112072812569176);
\fill[tum-red] (axis cs:1.86666666666667,0) rectangle (axis cs:2.13333333333333,0.0574113378289548);
\fill[tum-red] (axis cs:2.86666666666667,0) rectangle (axis cs:3.13333333333333,0.0468863492362365);
\addlegendimage{ybar,ybar legend,draw=tum-red,fill=tum-red}
\addlegendentry{ReLiCA* best}
\fill[tum-green] (axis cs:0.133333333333333,0) rectangle (axis cs:0.4,0.117086148982182);
\fill[tum-green] (axis cs:1.13333333333333,0) rectangle (axis cs:1.4,0.117367228900362);
\fill[tum-green] (axis cs:2.13333333333333,0) rectangle (axis cs:2.4,0.0616639837876511);
\fill[tum-green] (axis cs:3.13333333333333,0) rectangle (axis cs:3.4,0.0488770034062284);
\addlegendimage{ybar,ybar legend,draw=tum-green,fill=tum-green}
\addlegendentry{ReLiCA* worst}
\end{axis}

\end{tikzpicture}
    \caption{Comparison of the performance of the overall best linear rule with the best and worst rule selected by \gls{relicada}.}%
    \label{fig:barplot_reca}
\end{figure}
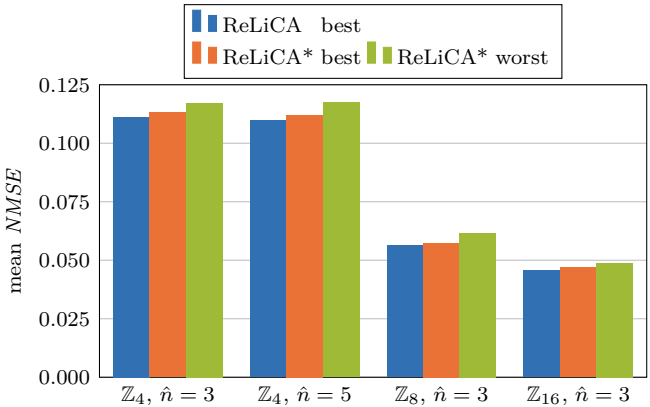

In \cref{fig:barplot_reca}, we compare the mean \(\nmse\) of the overall best \gls{relica} model, analyzing all possible linear rules, with the best and worst \gls{relica}* model, whose rules were selected by \gls{relicada}.
Best and worst are determined per dataset, resulting in the possibility that different rules are used for the different datasets.
It can be seen that the best \gls{relica}* model is very close to the overall best \gls{relica} model,
especially considering that the overall worst rule has a mean \(\nmse > 1\).
Not only the best \gls{relica}* model shows nearly optimal performance, but also the worst one.
It is also evident that increasing \(\hat{n}\) from 3 to 5 did not improve the performance.
This behavior was also verified for several other values of \(m\) (see \cref{sec:app_relicada}).

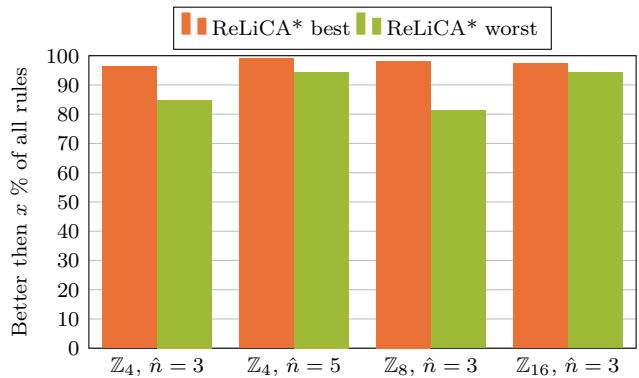
\begin{figure}[!t]
    \centering
    \begin{tikzpicture}
\begin{axis}[
axis background/.style={fill=white},
xmin=-0.5, xmax=3.5,
xtick={0,1,2,3},
ytick distance=10,
xticklabels={{\(\intnum_4\), \(\hat{n}=3\)},{\(\intnum_4\), \(\hat{n}=5\)},{\(\intnum_8\), \(\hat{n}=3\)},{\(\intnum_{16}\), \(\hat{n}=3\)}},
ylabel={Better then \(x\) \qty{}{\percent} of all rules},
ymajorgrids,
ymin=0.0, ymax=100.0,
width=\figurewidth,
height=\figureheight,
font=\footnotesize,
xtick style={draw=none},
ytick style={draw=none},
legend style={legend cell align=left, align=left, draw=black, font=\footnotesize, at={(0.5,1.02)},anchor=south},
legend columns=2,
]
\fill[tum-red] (axis cs:-0.4,0) rectangle (axis cs:0,96.2121212121212);
\fill[tum-red] (axis cs:0.6,0) rectangle (axis cs:1,98.967803030303);
\fill[tum-red] (axis cs:1.6,0) rectangle (axis cs:2,98.1601731601732);
\fill[tum-red] (axis cs:2.6,0) rectangle (axis cs:3,97.328431372549);
\addlegendimage{ybar,ybar legend,draw=tum-red,fill=tum-red}
\addlegendentry{ReLiCA* best}
\fill[tum-green] (axis cs:0,0) rectangle (axis cs:0.4,84.8484848484849);
\fill[tum-green] (axis cs:1,0) rectangle (axis cs:1.4,94.3939393939394);
\fill[tum-green] (axis cs:2,0) rectangle (axis cs:2.4,81.2409812409812);
\fill[tum-green] (axis cs:3,0) rectangle (axis cs:3.4,94.2179144385027);
\addlegendimage{ybar,ybar legend,draw=tum-green,fill=tum-green}
\addlegendentry{ReLiCA* worst}
\end{axis}

\end{tikzpicture}
    \caption{Rules selected by ReLiCADA are better than \(x\)\qty{}{\percent} of the overall linear rules.}%
    \label{fig:barplot_reca_percent}
\end{figure}
Instead of using only the mean \(\nmse\) for this analysis, we also checked how many \gls{relica} models are worse than a selected \gls{relica}* model.
The results are depicted in \cref{fig:barplot_reca_percent} and clearly show that the best \gls{relica}* model is at least better than \qty{95}{\percent} of the total rule space.
Even the worst \gls{relica}* model is still better than \qty{80}{\percent} of the overall \gls{relica} models.
This again verifies that the performance of all rules selected by \gls{relicada} is far better than randomly choosing a linear rule.

As it is reasonable to test all configurations selected by \gls{relicada} it is possible to always achieve the
best performance in \cref{fig:barplot_reca,fig:barplot_reca_percent}.

\begin{figure}[!t]
    \centering
    \begin{tikzpicture}
\begin{axis}[%
xmin=10000, xmax=10000000,
log basis x={10},
xmode=log,
xlabel={complexity},
ytick distance=0.05,
grid=both,
axis background/.style={fill=white},
legend style={legend cell align=left, align=left, draw=black, font=\footnotesize, at={(0.48,1.02)},anchor=south},
width=\figurewidth,
height=\figureheight,
font=\footnotesize,
ylabel={mean \(\nmse\)},
ymajorgrids,
ymin=0, ymax=0.275,
xtick style={draw=none},
ytick style={draw=none},
every y tick label/.append style={/pgf/number format/.cd,fixed,precision=2,fixed zerofill},
legend columns=4,
]

\addplot [
    scatter,
    only marks,
    point meta=explicit symbolic,
    scatter/classes={
        2b={mark=*,tum-red},
        3b={mark=diamond*,tum-red},
        4b={mark=square*,tum-red},
        5b={mark=pentagon*,tum-red},
        nn={mark=*,tum-yellow},
        scr={mark=*,tum-blue-brand},
        dlr={mark=diamond*,tum-blue-brand},
        esn={mark=square*,tum-blue-brand},
        lin={mark=*,tum-pink},
        gru={mark=*,tum-green},
        lstm={mark=diamond*,tum-green},
        rnn={mark=square*,tum-green}%
    },
] table [meta=label] { 
    x         y                      label
    218144    0.118758805340406      2b
    306208    0.0561645850427271     3b
    402464    0.0468863492362365     4b
    506912    0.0456404188162847     5b
    2689024   0.0744991413302956     dlr
    2690080   0.0989206875934393     scr
    4418112   0.103090096114326      nn
    4995072   0.130790649607432      rnn
    5299164   0.233645714052992      gru
    5932032   0.26283996626647       lstm
    5128192   0.0727653919730603     esn
    21120     0.141845145323623      lin
}; 
\legend{ReLiCA* \(\intnum_4\), ReLiCA* \(\intnum_8\), ReLiCA* \(\intnum_{16}\), ReLiCA* \(\intnum_{32}\), NN, SCR, DLR, ESN, Linear, GRU, LSTM, RNN}
\end{axis}
\end{tikzpicture}%
    \caption{Comparison of model performance with model complexity.}%
    \label{fig:complex}
\end{figure}
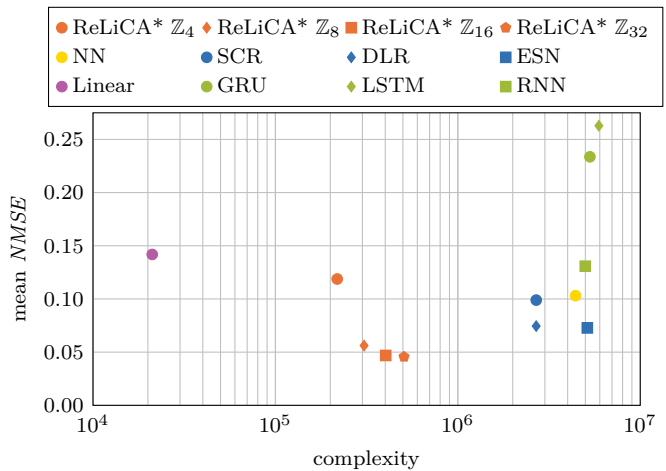
Since one goal was to achieve a computationally simple model with low complexity while maintaining good model performance, we compared these two parameters in \cref{fig:complex}.
We used a train-test-validation split of the dataset for this analysis.
The test performance values were used to select the best model, and the validation performance values are shown in \cref{fig:complex}.
No large deviations between test and validation performance were evident during our experiments.
The \gls{relica} models have less complexity compared to the \gls{rc} and \gls{nn} models.
Despite their computational simplicity, they still achieve similar or even better performance.
Increasing \(m\) for the \gls{relica} models increases not only the model complexity but also the model performance.
However, it is apparent that the performance gain by increasing \(m\) declines.
A neighborhood of \(\hat{n}=3\) was chosen for the \gls{relica} models since increasing the neighborhood would not result in better performance.

Despite the nonlinear, and thus more complex, \gls{ca} of the \textit{Babson} models, their performance is not up to
the \gls{relica}* models. While the \gls{relica}* \(\intnum_4\) models achieve a mean \nmse{} of \(0.12\), the
\textit{Babson} models only achieve \(0.34\). As the nonlinear CA rules of the Babson models have been optimized with a
\gls{ga}, this indicates that heuristic search and optimization algorithms cannot deal with the structure and size of the general \gls{ca} rule space very well.

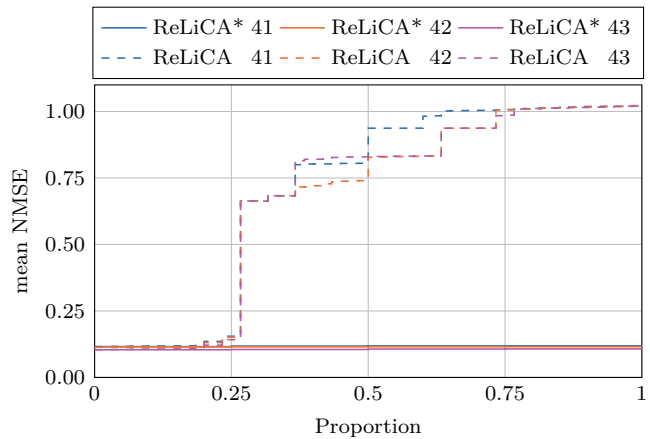
\begin{figure}[!t]
    \centering
    \begin{tikzpicture}
\begin{axis}[%
xlabel={Proportion},
xmin=0, xmax=1,
ytick distance=0.25,
xtick distance=0.25,
ymin=0, ymax=1.1,
ylabel={mean NMSE},
grid=major,
axis background/.style={fill=white},
legend style={legend cell align=left, align=left, draw=black, font=\footnotesize, at={(0.5,1.02)},anchor=south},
width=\figurewidth,
height=\figureheight,
font=\footnotesize,
xtick style={draw=none},
ytick style={draw=none},
legend columns=3,
every y tick label/.append style={/pgf/number format/.cd,fixed,precision=2,fixed zerofill},
scaled y ticks=false,
]
\addplot [semithick, tum-blue-brand, const plot mark right]
table {%
0.0 0.11581153142314271
0.25 0.11581153142314271
0.5 0.11777793106212141
0.75 0.11842111342523581
1.0 0.11872912849513578
};
\addlegendentry{ReLiCA* 41}
\addplot [semithick, tum-red, const plot mark right]
table {%
0.0 0.11499060146701828
0.25 0.11499060146701828
0.5 0.11503459326114183
0.75 0.11531625979669094
1.0 0.11551960457458663
};
\addlegendentry{ReLiCA* 42}
\addplot [semithick, tum-pink, const plot mark right]
table {%
0.0 0.10403772696920238
0.25 0.10403772696920238
0.5 0.10508145384098783
0.75 0.10681077691000214
1.0 0.107269888014762
};
\addlegendentry{ReLiCA* 43}
\addplot [semithick, tum-blue-brand, const plot mark right, dashed]
table {%
0.0 0.11581153142314271
0.016666666666666666 0.11581153142314271
0.03333333333333333 0.11587466329799777
0.05 0.11655464696157856
0.06666666666666667 0.11694639152675657
0.08333333333333333 0.11760301471655926
0.09999999999999999 0.11777793106212141
0.11666666666666667 0.1178192057566415
0.13333333333333333 0.11788022552556378
0.15 0.11808147116562394
0.16666666666666666 0.11842111342523581
0.18333333333333332 0.11872912849513578
0.19999999999999998 0.12003805330614165
0.21666666666666667 0.13526200401966745
0.23333333333333334 0.13526200407253575
0.25 0.15483718774110217
0.26666666666666666 0.15483718789656062
0.2833333333333333 0.6632529329493537
0.3 0.6632529342307063
0.31666666666666665 0.6632529342307063
0.3333333333333333 0.6830993780981826
0.35 0.6830993803260706
0.36666666666666664 0.6830993803260706
0.3833333333333333 0.8001842433965686
0.39999999999999997 0.8025892618666155
0.4166666666666667 0.8028999291852918
0.43333333333333335 0.8031316013060483
0.45 0.8043035612132357
0.4666666666666667 0.8047855868891698
0.48333333333333334 0.8050506254377964
0.5 0.8054223257506034
0.5166666666666667 0.9373444041198513
0.5333333333333333 0.9373444145314895
0.55 0.9373444145314895
0.5666666666666668 0.9373444145314895
0.5833333333333334 0.9373444407678381
0.6000000000000001 0.9373444407678381
0.6166666666666667 0.9821198344940697
0.6333333333333334 0.9836970501224322
0.65 1.0020004593298406
0.6666666666666667 1.0028946894865554
0.6833333333333333 1.0033415631735658
0.7000000000000001 1.0037028737890887
0.7166666666666667 1.0043163448575987
0.7333333333333334 1.0046525242686366
0.75 1.0055170898363488
0.7666666666666667 1.0064328561532025
0.7833333333333333 1.0085540151388352
0.8 1.0101457346778195
0.8166666666666668 1.0119963802915637
0.8333333333333334 1.0130171366336127
0.8500000000000001 1.0137598693999783
0.8666666666666667 1.0162184726715495
0.8833333333333334 1.0163406428971529
0.9 1.018754686932747
0.9166666666666667 1.0187591421444497
0.9333333333333333 1.0193279620346531
0.9500000000000001 1.019430254890217
0.9666666666666667 1.0195279230543721
0.9833333333333334 1.020654865937628
1.0 1.0212870873509914
};
\addlegendentry{ReLiCA\phantom{*} 41}
\addplot [semithick, tum-red, const plot mark right, dashed]
table {%
0.0 0.1136890166460976
0.016666666666666666 0.1136890166460976
0.03333333333333333 0.11427768451574159
0.05 0.1144721507428317
0.06666666666666667 0.11475050050947622
0.08333333333333333 0.11499060146701828
0.09999999999999999 0.11503459326114183
0.11666666666666667 0.1151387196414335
0.13333333333333333 0.11531625979669094
0.15 0.11551960457458663
0.16666666666666666 0.11557178270896305
0.18333333333333332 0.11584293696498839
0.19999999999999998 0.11637543763265104
0.21666666666666667 0.13095294792716217
0.23333333333333334 0.13095294935659155
0.25 0.15171233623236718
0.26666666666666666 0.1517123362955802
0.2833333333333333 0.6632529325632291
0.3 0.6632529342307063
0.31666666666666665 0.6632529343325747
0.3333333333333333 0.6830993782873809
0.35 0.6830993803260706
0.36666666666666664 0.6830993811386787
0.3833333333333333 0.716030322765243
0.39999999999999997 0.7167146805044798
0.4166666666666667 0.7208083686705126
0.43333333333333335 0.7279462556646407
0.45 0.7357239201714763
0.4666666666666667 0.7378641958065028
0.48333333333333334 0.739782076321791
0.5 0.7399636574666466
0.5166666666666667 0.8276039376336155
0.5333333333333333 0.8287120463654412
0.55 0.8304512998535208
0.5666666666666668 0.8305843676821656
0.5833333333333334 0.8309460639097783
0.6000000000000001 0.831219561357133
0.6166666666666667 0.832211629523324
0.6333333333333334 0.8329899004299335
0.65 0.9373444041198513
0.6666666666666667 0.9373444145314895
0.6833333333333333 0.9373444146478275
0.7000000000000001 0.9373444146478275
0.7166666666666667 0.9373444397456577
0.7333333333333334 0.9373444407678381
0.75 1.004601901560641
0.7666666666666667 1.0076374196308509
0.7833333333333333 1.0094213721971375
0.8 1.0096337337159538
0.8166666666666668 1.0099444423850934
0.8333333333333334 1.0128707073376568
0.8500000000000001 1.0138066206890457
0.8666666666666667 1.0150596884849994
0.8833333333333334 1.0156445813650856
0.9 1.016687081057809
0.9166666666666667 1.0168511881780997
0.9333333333333333 1.018018978462267
0.9500000000000001 1.018264670987219
0.9666666666666667 1.0184151149008374
0.9833333333333334 1.0187795786096767
1.0 1.022216866241582
};
\addlegendentry{ReLiCA\phantom{*} 42}
\addplot [semithick, tum-pink, const plot mark right, dashed]
table {%
0.0 0.10403772696920238
0.016666666666666666 0.10403772696920238
0.03333333333333333 0.10408464410788029
0.05 0.10508145384098783
0.06666666666666667 0.10527375986070409
0.08333333333333333 0.10660845707660214
0.09999999999999999 0.10669223964530943
0.11666666666666667 0.10681077691000214
0.13333333333333333 0.10693871226200678
0.15 0.10717424398092712
0.16666666666666666 0.107269888014762
0.18333333333333332 0.10786298521101832
0.19999999999999998 0.1086055999099783
0.21666666666666667 0.12141979943946207
0.23333333333333334 0.12141980275626865
0.25 0.1419723754285667
0.26666666666666666 0.14197237749756325
0.2833333333333333 0.6632529327595424
0.3 0.6632529342764544
0.31666666666666665 0.6632529342876275
0.3333333333333333 0.68309937855277
0.35 0.6830993797618332
0.36666666666666664 0.6830993809655287
0.3833333333333333 0.8146119197864305
0.39999999999999997 0.8202510019200482
0.4166666666666667 0.8205356720606432
0.43333333333333335 0.8219596264067895
0.45 0.827295044384308
0.4666666666666667 0.828818990145777
0.48333333333333334 0.8289504578896504
0.5 0.829278588429347
0.5166666666666667 0.8296835658798436
0.5333333333333333 0.8307971831986667
0.55 0.8316569765804424
0.5666666666666668 0.831706301543489
0.5833333333333334 0.8317791239306443
0.6000000000000001 0.8325180914022551
0.6166666666666667 0.8326803552588323
0.6333333333333334 0.8327420919570113
0.65 0.9373444041198513
0.6666666666666667 0.9373444145314895
0.6833333333333333 0.9373444145314895
0.7000000000000001 0.9373444145314895
0.7166666666666667 0.9373444407678381
0.7333333333333334 0.9373444407678381
0.75 0.984385942135788
0.7666666666666667 0.9864011267250639
0.7833333333333333 1.0085820988287966
0.8 1.011306177543669
0.8166666666666668 1.0114503417426972
0.8333333333333334 1.0127880876130186
0.8500000000000001 1.012794360556169
0.8666666666666667 1.0129085069867865
0.8833333333333334 1.0146655126160402
0.9 1.0149648403113454
0.9166666666666667 1.015555146248878
0.9333333333333333 1.016170022277227
0.9500000000000001 1.0173519885574043
0.9666666666666667 1.0197286693413783
0.9833333333333334 1.020627577075194
1.0 1.0212698555568474
};
\addlegendentry{ReLiCA\phantom{*} 43}
\end{axis}
\end{tikzpicture}%
    \caption{Influence of the random mapping seed on the ReLiCA model performance. The used model configuration is: \(\intnum_4\), \(\hat{n}=3\), \textit{scale\_offset}, \textit{replacement}.}%
    \label{fig:ecdf_seed}
\end{figure}

To analyze the influence of the random mapping on the \gls{relica} model performance, we tested several different seeds for the random mapping generator.
While there is an influence on the performance, it is neglectable for the models selected by \gls{relicada}.
In \cref{fig:ecdf_seed}, the empirical cumulative distribution function for different seeds is visualized for \(\intnum_4\), \(\hat{n}=3\) \gls{relica} and ReLiCA* models
using \textit{scale\_offset} and \textit{replacement}. The slight performance difference decreases even further with larger \(m\).

During our experiments, we mainly focused on the integer rings \(m=2^a\) with \(a \in \natnum^+\) since these are most suitable for implementations on \glspl{fpga} and other digital systems.
Nevertheless, we verified \gls{relicada} for several other values of \(m\) (see \cref{sec:app_relicada}). These results showed that \gls{relicada} can also be used for \(m \neq 2^a\).
According to our experiments, \glspl{ca} over \(\intnum_2\) behave differently.
For example, the best encoding for these \glspl{ca} is the \textit{xor} encoding.
Since this configuration was not of primary interest, we did not analyze this further.
Furthermore we verified \gls{relicada} on lattice sizes not equal to \({(16,32)}\) and iterations not equal to \(4\). Also, for these configurations, \gls{relicada} showed great improvements in performance compared to the whole set of all possible ReLiCA models. The performance values are listed in \cref{sec:app_relicada}.

We also ran tests where the quantized input \(x_q\) was directly fed into the readout layer, forming a quantized skip connection.
When the \textit{replacement} encoding was used, this did not lead to any performance gain.
Since \gls{relicada} only uses \textit{replacement} encoding, quantized skip-connections are not used in our models.
However, a performance gain was observed when the readout layer was provided with the original input \(x\) directly.
Since this imposes only a very little increase in complexity, we recommend using this skip connection if possible.

\subsection{Nonlinear Capabilities}%
\label{subsec:nonlinear_capabilities}

During our experiments, we saw that \gls{relica} models could not deal with highly nonlinear datasets, like \textit{H\'{e}non}, very well.
However, after using the hyperparameter optimizer \texttt{Optuna} to optimize the quantization thresholds (see \cref{eq:quantization}) and the regularization of the Ridge optimizer, the performance of the \gls{relica} model increased drastically.
The \gls{relica}* model with \(\intnum_{16}\), \(\hat{n}=3\) achieved an \nmse{} of \(0.321\) before optimization and \(0.048\) after.
Other transformation and quantization layers could likely improve the nonlinear capabilities of linear \gls{relica} models. However, this was not further analyzed.

Further tests have shown that the \gls{relica} model performance also improves on the other datasets when \texttt{Optuna} is used to optimize quantization thresholds.
Since we wanted to create a fast and easy-to-train model, we refrained from using threshold optimization in our results.
\section{Conclusion}%
\label{sec:conclusion}

\gls{reca} represents a particular form of the broader field of \gls{rc} that is particularly suited to be implemented on \glspl{fpga}.
However, the choice of hyperparameters
and, primarily, the search for suitable \gls{ca} rules are major challenges during the design phase of such models.
When restricted to linear \glspl{ca}, fundamental properties can be computed analytically.
Based on the results of nearly a million experiments, we recognized that linear \gls{ca} rules that achieve low errors on many relevant benchmark datasets have specific mathematical properties.
Based on these insights, we developed the \acrlong{relicada}, which selects hyperparameters that have been shown to work well in the experiments.
Most importantly, the proposed algorithm pre-selects a few rules out of the rule space that grows exponentially with increasing \(m\) and \(\hat{n}\).
As it has been shown, the best-performing selected rules are among the top \(\qty{5}\percent\) of the overall rule space.
Moreover, the proposed models achieve, on average, a lower error than other state-of-the-art neural network models and, at the same time, exhibit less computational complexity, showing the strength of \gls{relicada}.
Furthermore, with the immensely reduced hyperparameter space, the time needed to design and implement \gls{reca} models is drastically reduced.
In conclusion, \gls{relicada} is a promising approach for designing and implementing \gls{reca} models for time series processing and analysis.
{\appendices%
\crefalias{section}{appsec}
\section{Performance Values of Compared Models}%
\label{sec:app_baseline}
\Cref{tab:baseline,tab:reca_perf1,tab:reca_perf2} list the \nmse{} values of the models compared throughout this paper.
The dark blue color highlights the model with the lowest \nmse{} for the respetive dataset.
The light blue color indicates that the model has a similar performance (same value rounded to 3 decimal places) compared to the best model.

The validation performance values of the reference models are listed in \cref{tab:baseline}. All models except the Linear model were optimized using Optuna.

\Cref{tab:reca_perf1,tab:reca_perf2} list the performance values of the \gls{reca} models for the test and validation set.

\section{Supplementary Materials}
Supplementary files to this paper can be found in the git repository at \url{https://github.com/jkantic/ReLiCADA}.

\section{Performance Values of ReLiCADA}%
\label{sec:app_relicada}
The \nmse{} values of all tested \gls{relicada} models are listed in the file \textit{relicada\_performance.xlsx} (supplementary materials).
The table lists the number of rules selected by \gls{relicada} as well as the minimum, mean, and maximum \nmse{} values of the selected rules.
If the whole rule space for a given configuration was tested, these values are also calculated for the whole set of rules. Furthermore, the percentage of rules worse than the best/worst \gls{relicada} rule is stated.

\section{Tested Configurations}
The file \textit{configs.xlsx} (supplementary materials) contains all \gls{relica} model configurations tested.
The trans/quant column lists the used transformation and quantization algorithm.

\section{Raw Experiment Output}
The supplementary materials to this paper include CSV files containing the raw experiment results for all tested configurations.
For each dataset, a separate CSV file lists: \((N_r,N_c)\), \(I\), \(m\), \(\hat{n}\), transformation, quantization,
encoding, mapping, seed, \(\vec{w}\), and the \nmse{} performance value.

\begin{table*}
    \caption{Baseline Model Performance}%
    \label{tab:baseline}
    \centering
    \begin{tblr}{hlines, colspec=|c|[1pt]c|c|c|c|c|c|c|c|,
                  hline{1} = {1}{0pt},
                  vline{1} = {1}{0pt},}
          & NN & RNN & GRU & LSTM & DLR & SCR & ESN & Linear \\\SetHline[1]{2-9}{1pt}
        MG & \SetCell{tum-blue-light-2} 0.000 & 0.001 & 0.001 & 0.001 & \SetCell{tum-blue-light-2} 0.000 & \SetCell{tum-blue-light-2} 0.000 & \SetCell{tum-blue-light-2} 0.000 & \SetCell{tum-blue-light} 0.000\\
        MG\_25 & \SetCell{tum-blue-light-2} 0.000 & 0.099 & 0.030 & 0.015 & 0.001 & 0.001 & \SetCell{tum-blue-light} 0.000 & 0.310\\
        MSO & 0.001 & 0.074 & 0.037 & 0.092 & 0.002 & 0.001 & 0.002 & \SetCell{tum-blue-light} 0.000\\
        MSO\_3 & 0.004 & 0.414 & 0.665 & 0.723 & 0.040 & 0.096 & 0.079 & \SetCell{tum-blue-light} 0.000\\
        NARMA\_10 & \SetCell{tum-blue-light} 0.021 & 0.107 & 0.627 & 0.403 & 0.062 & 0.055 & 0.116 & 0.161\\
        NARMA\_20 & 0.551 & 0.165 & 0.552 & 0.544 & 0.161 & 0.439 & \SetCell{tum-blue-light} 0.124 & 0.542\\
        NARMA\_30 & 0.544 & \SetCell{tum-blue-light} 0.449 & 0.542 & 0.538 & 0.545 & 0.488 & 0.476 & 0.536\\
        NCC & \SetCell{tum-blue-light} 0.002 & 0.018 & 0.033 & 0.455 & 0.007 & 0.006 & \SetCell{tum-blue-light-2} 0.002 & 0.010\\
        PMP & \SetCell{tum-blue-light-2} 0.000 & \SetCell{tum-blue-light-2} 0.000 & \SetCell{tum-blue-light-2} 0.000 & 0.001 & \SetCell{tum-blue-light-2} 0.000 & \SetCell{tum-blue-light-2} 0.000 & \SetCell{tum-blue-light-2} 0.000 & \SetCell{tum-blue-light} 0.000\\
        PPST & \SetCell{tum-blue-light-2} 0.000 & 0.002 & 0.001 & 0.022 & \SetCell{tum-blue-light-2} 0.000 & \SetCell{tum-blue-light-2} 0.000 & \SetCell{tum-blue-light-2} 0.000 & \SetCell{tum-blue-light} 0.000\\
        PPST\_10 & 0.010 & 0.109 & 0.084 & 0.097 & 0.002 & 0.002 & 0.002 & \SetCell{tum-blue-light} 0.000\\
    \end{tblr}
\end{table*}

\begin{table*}
  \caption{\Gls{reca} Model Performance (Test-Set)}%
  \label{tab:reca_perf1}
  \centering
  \begin{tblr}{hlines, colspec=|c|[1pt]c|c|c|c|c|c|c|c|,
                cell{1}{1} = {r=2}{c},
                hline{1} = {1}{0pt},
                vline{1} = {1-2}{0pt},}
        & Babson & \SetCell[c=7]{c} ReLiCADA \\
        & {\(\intnum_4\) \\ \(\hat{n}=3\)} & {\(\intnum_4\) \\ \(\hat{n}=3\)} & {\(\intnum_4\) \\ \(\hat{n}=5\)} & {\(\intnum_8\) \\ \(\hat{n}=3\)} & {\(\intnum_8\) \\ \(\hat{n}=5\)} & {\(\intnum_{16}\) \\ \(\hat{n}=3\)} & {\(\intnum_{16}\) \\ \(\hat{n}=5\)} & {\(\intnum_{32}\) \\ \(\hat{n}=3\)}\\\SetHline[1]{2-16}{1pt}
      MG & 0.033 & 0.005 & 0.005 & 0.002 & \SetCell{tum-blue-light} 0.001 & 0.002 & 0.002 & 0.002\\
      MG\_25 & 0.052 & 0.018 & 0.013 & \SetCell{tum-blue-light-2} 0.004 & \SetCell{tum-blue-light} 0.004 & 0.008 & 0.008 & 0.034\\
      MSO & 0.510 & 0.112 & 0.113 & 0.029 & 0.029 & 0.008 & 0.008 & \SetCell{tum-blue-light} 0.002\\
      MSO\_3 & 0.647 & 0.129 & 0.129 & 0.033 & 0.031 & 0.009 & 0.009 & \SetCell{tum-blue-light} 0.002\\
      NARMA\_10 & 0.621 & 0.238 & 0.236 & 0.176 & 0.173 & 0.166 & 0.163 & \SetCell{tum-blue-light} 0.160\\
      NARMA\_20 & 0.649 & 0.204 & 0.204 & 0.145 & 0.145 & 0.134 & 0.135 & \SetCell{tum-blue-light} 0.131\\
      NARMA\_30 & 0.743 & 0.228 & 0.228 & 0.166 & 0.167 & 0.154 & 0.155 & \SetCell{tum-blue-light} 0.153\\
      NCC & 0.367 & 0.102 & 0.103 & 0.025 & 0.026 & 0.012 & 0.013 & \SetCell{tum-blue-light} 0.009\\
      PMP & 0.002 & \SetCell{tum-blue-light-2} 0.000 & \SetCell{tum-blue-light-2} 0.000 & \SetCell{tum-blue-light-2} 0.000 & \SetCell{tum-blue-light-2} 0.000 & \SetCell{tum-blue-light-2} 0.000 & \SetCell{tum-blue-light-2} 0.000 & \SetCell{tum-blue-light} 0.000\\
      PPST & 0.050 & 0.066 & 0.066 & 0.011 & 0.012 & 0.003 & 0.003 & \SetCell{tum-blue-light} 0.001\\
      PPST\_10 & 0.104 & 0.141 & 0.136 & 0.040 & 0.043 & 0.019 & 0.018 & \SetCell{tum-blue-light} 0.005\\
  \end{tblr}
\end{table*}

\begin{table*}
    \caption{\Gls{relicada} Model Performance (Validation-Set)}%
    \label{tab:reca_perf2}
    \centering
    \begin{tblr}{hlines, colspec=|c|[1pt]c|c|c|c|c|c|c|,
                  hline{1} = {1}{0pt},
                  vline{1} = {1}{0pt},}
          & {\(\intnum_4\) \\ \(\hat{n}=3\)} & {\(\intnum_4\) \\ \(\hat{n}=5\)} & {\(\intnum_8\) \\ \(\hat{n}=3\)} & {\(\intnum_8\) \\ \(\hat{n}=5\)} & {\(\intnum_{16}\) \\ \(\hat{n}=3\)} & {\(\intnum_{16}\) \\ \(\hat{n}=5\)} & {\(\intnum_{32}\) \\ \(\hat{n}=3\)}\\\SetHline[1]{2-16}{1pt}
        MG & 0.006 & 0.005 & 0.002 & \SetCell{tum-blue-light} 0.001 & 0.002 & 0.002 & 0.002\\
        MG\_25 & 0.018 & 0.016 & \SetCell{tum-blue-light} 0.004 & \SetCell{tum-blue-light-2} 0.004 & 0.010 & 0.010 & 0.036\\
        MSO & 0.115 & 0.120 & 0.031 & 0.030 & 0.008 & 0.008 & \SetCell{tum-blue-light} 0.002\\
        MSO\_3 & 0.137 & 0.138 & 0.034 & 0.034 & 0.009 & 0.010 & \SetCell{tum-blue-light} 0.002\\
        NARMA\_10 & 0.236 & 0.235 & 0.175 & 0.175 & 0.166 & \SetCell{tum-blue-light} 0.163 & 0.165\\
        NARMA\_20 & 0.179 & 0.182 & 0.130 & 0.130 & 0.120 & 0.119 & \SetCell{tum-blue-light} 0.118\\
        NARMA\_30 & 0.209 & 0.211 & 0.161 & 0.163 & 0.153 & \SetCell{tum-blue-light} 0.150 & 0.151\\
        NCC & 0.102 & 0.100 & 0.028 & 0.027 & 0.011 & 0.011 & \SetCell{tum-blue-light} 0.009\\
        PMP & \SetCell{tum-blue-light-2} 0.000 & \SetCell{tum-blue-light-2} 0.000 & \SetCell{tum-blue-light-2} 0.000 & \SetCell{tum-blue-light-2} 0.000 & \SetCell{tum-blue-light-2} 0.000 & \SetCell{tum-blue-light} 0.000 & \SetCell{tum-blue-light-2} 0.000\\
        PPST & 0.111 & 0.107 & 0.014 & 0.013 & 0.003 & 0.003 & \SetCell{tum-blue-light} 0.001\\
        PPST\_10 & 0.195 & 0.205 & 0.039 & 0.039 & 0.020 & 0.022 & \SetCell{tum-blue-light} 0.005\\
    \end{tblr}
\end{table*}

\section{ReLiCADA Pseudocode}%
\label{sec:pseudocode}

The pseudocode in \cref{code:relicada-rule-selection} together with \cref{eq:alg-allways,eq:alg-ring,eq:alg-field,eq:alg-two-entropy}
and the explanation in \cref{subsec:relicada} can be used to implement \gls{relicada}.

\begin{algorithm}
    \begin{algorithmic}[1]
    \Require{}\(N, m, \hat{n}\)
    \Ensure{}\(\text{Selected rules } S\)
    \State{}\(S \gets \emptyset\) %
    \State{}\(A \gets \emptyset\) %
    \State{}\(\primeset, \set{K} \gets \Func{prime\_factors}(m)\) \Comment{see \cref{eq:prime_set,eq:prime_multiplicities}}
    \State{}\(\primeset_w, \bar{\primeset}_w \gets \Func{split\_ring}(m)\) \Comment{see \cref{eq:prime_weights,eq:non_prime_weights}}
    \ForEach{\(\vec{w} = (w_{-r},\ldots,w_r) \in \mathcal{R}(m, \hat{n})\)} \Comment{see \cref{eq:rule_space}}
        \State{}\(A \gets A \bigcup \{\vec{w}\}\) %
        \State{}\(\hat{\vec{w}} = \Func{mirror}(\vec{w})\)

        \If{\(\hat{\vec{w}} \in A\)}
            \LeftComment{see \cref{eq:alg-rev}}
            \State{}\Continue%
        \EndIf%

        \State{}\(\nentropy \gets \Func{normalized\_entropy}(\primeset, \set{K}, \vec{w})\) \Comment{see \cref{eq:normalized_entropy}}
        \If{\(\nentropy \ne 1\)}
            \LeftComment{see \cref{eq:alg-nentropy}}
            \State{}\Continue%
        \EndIf%

        \ForEach{\(p \in \primeset\)}
            \State{}\(P \gets \{i : p \nmid w_i\}\)
            \If{\(\lvert P \rvert \ne 1\)}
                \LeftComment{see \cref{eq:alg-injective}}
                \State{}\Continue%
            \EndIf%
        \EndFor%

        \If{\(\Func{is\_prime}(m)\)} \Comment{\(\intnum_m\) is a field}
            \If{\(\exists w_i: (w_i \ne 0) \land (\subgroup^{\times}(w_i) \ne \intnum_m {\setminus} 0)\)}
                \LeftComment{see \cref{eq:alg-field}}
                \State{}\Continue%
            \EndIf%

        \ElsIf{\(\vert\primeset\vert = 2\)} \Comment{\(\intnum_m\) is a ring, \(m=p_1^{k_1}p_2^{k_2}\)}
            \State{}\(\{p_1,p_2\} = \primeset\)
            \If{\((\forall w_i,w_j): (p_1 \mid w_i) \lor (p_2 \mid w_j)\)}
                \LeftComment{see \cref{eq:alg-two-entropy}}
                \State{}\Continue%
            \EndIf%

        \Else{}\Comment{\(\intnum_m\) is a ring}
            \State{}\(N \gets \{i : w_i \ne 0\}\)
            \If{\(\lvert N \rvert \ne 2\)}
                \LeftComment{see \cref{eq:alg-zero-weights}}
                \State{}\Continue%
            \EndIf%
            \If{\(\exists w_i: (w_i \in \primeset_w) \land (w_i \notin \{1, m-1\})\)}
                \LeftComment{see \cref{eq:alg-prime-group}}
                \State{}\Continue%
            \EndIf%
            \If{\(m \ne 4\)}
                \If{\(\exists w_i: (w_i \in \bar{\primeset}_w) \land (\lvert \subgroup^{+}(w_i) \rvert \ne 4)\)}
                    \LeftComment{see \cref{eq:alg-subgroup}}
                    \State{}\Continue%
                \EndIf%
            \EndIf%
        \EndIf%
        \State{}\(S \gets S \bigcup \{\vec{w}\}\)
    \EndFor%
    \State{}\Return{}\(S\)
\end{algorithmic}

    \caption{ReLiCADA - Rule Selection}%
    \label{code:relicada-rule-selection}
\end{algorithm}
}

\bibliographystyle{IEEEtran}
\bibliography{tex/IEEEabrv, tex/Promotion.bib}

\vfill

\end{document}